\documentclass{article}

 \usepackage[eandd, final]{neurips_2026}

\usepackage[utf8]{inputenc} 
\usepackage[T1]{fontenc}    
\usepackage{hyperref}       
\usepackage{xurl}            
\usepackage{booktabs}       
\usepackage{amsfonts}       
\usepackage{nicefrac}       
\usepackage{microtype}      
\usepackage{xcolor}         
\usepackage{amsmath} 
\usepackage{graphicx}
\usepackage{subcaption}
\usepackage{changepage,threeparttable}
\usepackage{siunitx}
\title{Auditing Cross-Lingual Fairness in Language Model Watermarking}

\author{
  {\bf Alexander Nemecek, Osama Zafar, Debargha Ganguly, Vikash Singh,} \\
  {\bf Vipin Chaudhary, Erman Ayday} \\
  Case Western Reserve University \\
  \texttt{\{ajn98,oxz23,dxg512,vxs465,vxc204,exa208\}@case.edu}
}

\begin{document}

\maketitle

\begin{abstract}
Watermarking schemes for large language model output are evaluated almost exclusively on English text using each scheme's detection threshold and a narrow set of quality measurements. Multilingual deployment exposes evaluation-design choices that are inconsequential on English but determine conclusions cross-lingually. We propose an evaluation framework with four components: detection thresholds calibrated empirically per deployment context, a threshold-independent companion measurement that distinguishes calibration failures from detection failures, three disjoint quality measurement paradigms (distributional, paired-semantic, and reference-perplexity), and a generalized-entropy decomposition of cross-language disparity over a typological family partition. Applied to six watermarking schemes, three open-weight generators, eleven languages spanning four scripts and eight typological families, and both base and instruction-tuned regimes, the framework reveals failure modes that single-language, single-paradigm evaluation cannot surface. Against a shape-matched null, detection disparity is predominantly between-family on the typological partition, indicating that detection gaps are structural to language properties rather than idiosyncratic to particular languages; quality disparity is large but not reliably family-structured. 
\end{abstract}

\section{Introduction}\label{intro}
Recent advances in large language models (LLMs) have made AI-generated text difficult to distinguish from human-authored writing. This raises concerns about misinformation, the erosion of trust in online content, and model collapse, in which the performance of models trained on partially machine-generated data degrades over successive generations~\cite{chen2024combating, spitale2023ai, shumailov2024ai}. The misinformation concern in particular is not hypothetical. NewsGuard has identified over 3{,}000 AI-generated news and information sites publishing in 16 languages, often under legitimate-sounding names and with little to no human oversight~\cite{newsguard_ai_tracking_center}.

Text watermarking offers one mitigation by embedding a recoverable statistical signature into model output. A watermarking scheme modifies the generation process in a key-dependent way, most commonly by biasing the next-token distribution toward a pseudorandomly selected vocabulary subset~\cite{kirchenbauer2023watermark, nemecek-etal-2026-topic, dathathri2024scalable}. A paired detector then tests a candidate text against this signature and returns a score, which is thresholded to yield a detection decision~\cite{he2026optimal}. Schemes are typically evaluated on detectability, generation quality (fluency, coherence, and distributional similarity to unwatermarked output), and robustness to downstream perturbations~\cite{piet2025markmywords, tu2024waterbench, liang2024watermark}.

These evaluations, however, are conducted almost exclusively on English prompts and English generated text. The implicit assumption is that a scheme which detects reliably and preserves quality on English will do the same on other languages, but this assumption has not been empirically tested at scale. Our concern is with this English-centric evaluation practice, not with any presumption that English users come out ahead; the audit lets the data identify which languages are underserved in each configuration (\S\ref{sec:framework}). This gap matters in practice, because watermarking is already deployed in settings where users prompt and generate in dozens of languages~\cite{dathathri2024scalable, nemecek2025watermarking}, and the multilingual misinformation ecosystem flagged above is precisely the threat surface that motivates the technology in the first place.

\textbf{Our work.} We present the first systematic audit of cross-lingual fairness in LLM text watermarking. In plain terms, we ask whether a watermark gives speakers of every language the same protection against AI-generated text and imposes the same quality cost, the same question of equal treatment across groups that has been asked of language technology broadly~\cite{blasi-etal-2022-systematic} and of classifiers in fair-ML practice~\cite{eeoc1978uniform, speicher2018unified}. Our findings answer no, and the gaps for detection line up with whole language families rather than with individual languages. Our contribution is methodological and empirical: an evaluation framework that treats multilingual fairness as a first-class object of measurement rather than an after-the-fact slice of an English-centric benchmark, and a characterization of how current schemes behave under it.

\textbf{-\quad \textit{Framework.}} We propose four evaluation components designed for cross-lingual settings: empirically calibrated detection thresholds, a threshold-independent companion measurement that distinguishes calibration failures from detection failures, three disjoint quality paradigms (distributional, paired-semantic, and reference-perplexity), and a generalized-entropy decomposition of disparity over a pre-registered typological partition.

\textbf{-\quad \textit{Audit.}} We apply the framework to six watermarking schemes, three open-weight generators, eleven languages spanning four scripts and eight typological families, and two generation regimes (parallel continuation and native-speaker instruction), yielding $\approx200{,}000$ matched watermarked/unwatermarked generations on shared prompts.

\textbf{-\quad \textit{Findings.}} Cross-language disparity is large and scheme-specific; on detection it is predominantly between-family rather than idiosyncratic to particular languages, and on quality it is not reliably family-structured. Per-scheme rankings depend on which evaluation choice (threshold, paradigm, prompt regime) is taken as canonical, and we identify failure modes invisible to single-language, single-paradigm evaluation.

\section{Related Work}\label{related}

\textbf{Watermarking schemes.} Two design families organize recent work on LLM text watermarking. The green-list logit-bias family biases the next-token distribution toward a pseudorandomly selected vocabulary subset, with variants differing in how the partition is constructed: per-position from a sliding window of preceding tokens~\cite{kirchenbauer2023watermark}, statically across the entire vocabulary~\cite{zhao2023provable}, or via a cross-lingual semantic-cluster lookup intended to preserve detectability under translation~\cite{he2024can}. The distortion-free family aims to leave the marginal next-token distribution unchanged in expectation, intervening at sampling time through permutation reparameterization~\cite{wu2023resilient}, key-conditioned tournament sampling~\cite{dathathri2024scalable}, or inverse transform sampling against a pseudo-random uniform stream~\cite{kuditipudi2023robust}.

\textbf{Watermark evaluation.} Existing watermark evaluation frameworks share a methodological template: detection is reported at each scheme's default threshold targeting a theoretical false-positive rate, quality is reported under a single paradigm, and prompts and generations are English. MarkMyWords~\cite{piet2025markmywords} and WaterBench~\cite{tu2024waterbench} instantiate this template at benchmark scale, and dedicated robustness studies extend it to adversarial perturbations while retaining the same threshold and quality conventions~\cite{liang2024watermark}. These choices are inconsequential on English under single-paradigm reporting, and become consequential cross-lingually. Recent work has identified this gap, calling for cross-lingual and demographic disaggregation~\cite{nemecek2026gets}. Prior cross-lingual evaluation of watermarking has addressed translation robustness, asking whether a signal embedded in one language survives a translation pass, rather than whether detection and quality are preserved across languages of original generation~\cite{he2024can}.

\textbf{Fairness methodology.} A watermark detector is a binary classifier and a quality measurement is a real-valued performance score, so per-language detection and quality vectors are the watermarking analog of per-group performance vectors in classical fair-ML evaluation. We import three tools from that literature. Operating-point calibration sets detection thresholds per group rather than globally to expose calibration-driven disparity~\cite{hardt2016equality}. The four-fifths rule flags disparate impact when a group scores below 80\% of the best group~\cite{feldman2015certifying}. Generalized-entropy decomposition splits a measure of spread across groups into within- and between-partition parts under a partition fixed in advance~\cite{shorrocks1980class}.

\section{Experimental Grid}\label{sec:grid}
We evaluate every cell of an $11 \text{ languages} \times 6 \text{ schemes} \times 3 \text{ generators} \times 2 \text{ regimes}$ grid at $n = 500$ matched pairs per cell (one watermarked, one unwatermarked, on a shared prompt). The grid's axes are chosen so that the typological decomposition of \S\ref{subsec:ge} is well-posed (multiple languages per script and morphological class), so that base-versus-instruct sensitivity is identifiable (every generator appears in both regimes), and so that scheme-family effects are not confounded with any single tokenizer (every scheme is run on three independent vocabularies).

\textbf{Languages:} The eleven evaluation languages span four scripts, eight typological families, and two Joshi et al. resource tiers~\cite{joshi2020state}; Table~\ref{tab:languages} enumerates the assignments. The set is the minimal configuration in which the between-family component of the generalized-entropy decomposition (\S\ref{subsec:ge}) is identifiable on more than one non-singleton family: Germanic and Romance both carry within-family signal, and the remaining six families enter as singletons, whose findings we attribute to the sampled language (Turkish rather than Turkic).

\begin{table}[htbp]
\centering
\scriptsize
\begin{tabular}{llllc}
\toprule
ISO 639-3 & Language & Script & Typological family & Joshi et al. tier \\
\midrule
eng & English      & Latin       & Germanic       & 5 \\
nld & Dutch        & Latin       & Germanic       & 4 \\
fra & French       & Latin       & Romance        & 5 \\
spa & Spanish      & Latin       & Romance        & 5 \\
por & Portuguese   & Latin       & Romance        & 4 \\
tur & Turkish      & Latin       & Turkic         & 4 \\
vie & Vietnamese   & Latin       & Austroasiatic  & 4 \\
hin & Hindi        & Devanagari  & Indic          & 4 \\
arb & Arabic & Arabic      & Semitic        & 5 \\
zho & Chinese      & Han         & Sinitic        & 5 \\
jpn & Japanese     & Han + Kana  & Japonic        & 5 \\
\bottomrule
\end{tabular}
\caption{The eleven evaluation languages with script, typological family, and Joshi resource tier.}
\label{tab:languages}
\end{table}

\textbf{Watermarking schemes:} We evaluate six schemes covering the two design families introduced in \S\ref{related}. The green-list logit-bias family contributes KGW~\cite{kirchenbauer2023watermark} (per-position sliding window), Unigram~\cite{zhao2023provable} (static green list), and XSIR~\cite{he2024can} (cross-lingual semantic-cluster lookup, designed to preserve detectability under translation). The distortion-free family contributes DIP~\cite{wu2023resilient} (permutation reparameterization), SynthID-Text~\cite{dathathri2024scalable} (key-conditioned tournament sampling), and EXPEdit~\cite{kuditipudi2023robust} (inverse transform sampling). For EXPEdit we use the published fast-path detector, which replaces the original permutation test with a closed-form score. Together the six schemes vary along the axes whose interaction with cross-lingual generation is the object of study: per-position versus static green-list construction, logit-bias versus sampling-time intervention, and English-trained versus cross-lingual partitioning.

\textbf{Generators and regimes:} We use three open-weight generator families with distinct tokenizers and meaningful coverage of our eleven languages: Mistral-NeMo-12B~\cite{mistralai2024nemo}, Gemma-3-4B~\cite{gemma3_2025}, and Qwen2.5-7B~\cite{yang2024qwen2}. Each is run in both its base and instruction-tuned variant. The two regimes control different sources of cross-lingual confound. The \textbf{base regime} feeds each base generator a FLORES$+$ devtest sentence~\cite{nllb-24} as a continuation prompt; FLORES$+$ provides parallel translations of the same source sentences across all eleven languages, so prompt content is held constant across languages and any cross-lingual variation in detection or quality cannot be attributed to prompt-content variation. The \textbf{instruct regime} feeds each instruction-tuned generator native-speaker prompts from the Aya dataset~\cite{singh2024aya} (\texttt{original-annotations} subset only, no machine translations); Aya provides linguistically natural prompt distributions written by speakers of the target language, so cross-lingual variation reflects realistic deployment input. 

\textbf{Generation protocol and LID gating:}
For every cell we generate $n = 500$ watermarked outputs and an additional $n = 500$ unwatermarked outputs sharing the same prompts and seeds, fixing $\texttt{temperature} = 0.7$, $\texttt{max\_new\_tokens} = 200$, and $\texttt{min\_new\_tokens} = 100$. The unwatermarked baseline is shared across the six schemes within each $(l, m, r)$ triple, so detection comparisons within a cell are matched-prompt. Every generation passes through GlotLID-v3~\cite{kargaran2023glotlid} for language identification (LID), and text in the prompted language counts as on-target. The headline numbers in \S\ref{sec:findings} use \texttt{subset = all} (no LID filtering) to avoid post-hoc selection effects, with on-target and off-target slices in Appendices~\ref{app:detection_threshold_robustness} and~\ref{app:qstratify}. Cells with fewer than 200 LID-passing unwatermarked generations are excluded from quantile calibration at $\alpha = 0.01$ (100 at $\alpha = 0.05$); paired-set quality measurements require at least 100 samples per side.

\section{Evaluation Framework}\label{sec:framework}
We propose a four-component framework: empirically calibrated detection thresholds (\S\ref{subsec:tau}), a threshold-independent check that tells a misplaced threshold from a missing signal (\S\ref{subsec:auc}), three quality measurements (\S\ref{subsec:quality}), and a split of disparity into between- and within-family parts (\S\ref{subsec:ge}). \S\ref{subsec:m1m4} lists the summary measurements (M1-M4) and Appendix~\ref{app:framework-details} collects the formal definitions.

\textbf{Assumptions and choices.} Robustness evaluation asks whether the signal survives changes made to a text. We ask instead whether a watermark's protection and its quality cost are shared equally across languages, that is, whether AI-generated text is caught at the same rate in every language and whether text is damaged equally little. Each measure in \S\ref{subsec:m1m4} answers one plain question. M1 shows every language directly, M2 gives the mean and the worst-served language, M3 counts the languages below 80\% of the best, and M4 asks whether the gaps line up with whole families or scatter language by language.

The four-fifths rule anchors to the best-served group rather than to English~\cite{eeoc1978uniform, feldman2015certifying}, so the advantaged group is identified from the data rather than assumed, and varies by cell (XSIR-Qwen base: Hindi TPR 0.82, English 0.17). We use generalized entropy because it is the class of inequality measures that splits additively into within- and between-family parts~\cite{shorrocks1980class}, a property since adopted in algorithmic fairness~\cite{speicher2018unified}. Max-min gaps do not split, and variance changes if the scores are rescaled. ``Underserved'' here means worst measured performance in a given configuration, narrower than the socially grounded notion in the fairness literature~\cite{barocas2023fairness, selbst2019fairness}. The components need only a per-item score, a group label, and a partition fixed beforehand, so the same audit applies to demographic or dialect groups~\cite{nemecek2026gets}.

\subsection{Empirical FPR Threshold Calibration}
\label{subsec:tau}
We index each evaluation cell as $c = (l, s, m, r)$ for language, scheme, generator and regime, write $S_s$ for scheme $s$'s detection score, which is higher on watermarked text, and $\mathcal{X}^+_c$ and $\mathcal{X}^-_c$ for the watermarked and matched-prompt unwatermarked sets in $c$. The schemes' shipped thresholds assume a null of independently drawn tokens, which cross-lingual generation does not satisfy, so we calibrate empirically instead: at level $\alpha \in (0, 1)$,
\[
\tau_c(\alpha) \;:=\; \mathrm{Quantile}_{1-\alpha}\!\left(\{S_s(x) : x \in \mathcal{X}^-_c\},\ \mathrm{lower}\right),
\]
the largest cutoff that keeps the false-positive rate on $\mathcal{X}^-_c$ at or below $\alpha$. A per-language threshold $\tau^{\mathrm{lang}}_c(\alpha)$ models a separate detector for each language; the global threshold $\tau^{\mathrm{global}}_{(s,m,r)}(\alpha)$, set on the pooled unwatermarked text of all languages, models the one shared detector a deployment uses and is our headline. The pair is the cross-lingual form of operating-point calibration, the per-group threshold setting of fair classification~\cite{hardt2016equality}.

\subsection{Threshold-Independent Companion Measurement}\label{subsec:auc}
The $\mathrm{TPR}$ at $\tau_c(\alpha)$ can collapse to zero when unwatermarked scores pile up at the top of the score scale and push the threshold there, even though the two distributions are otherwise well separated. The threshold-independent companion is the probability that a watermarked text outscores an unwatermarked one,
\[
\mathrm{AUC}_c \;=\; \Pr\!\left(S_s(x^+) > S_s(x^-)\right), \quad x^+\!\in\!\mathcal{X}^+_c,\ x^-\!\in\!\mathcal{X}^-_c.
\]
$\mathrm{TPR}$ near zero with $\mathrm{AUC}$ well above 0.5 marks a calibration failure, and with $\mathrm{AUC}$ near 0.5 a detection failure.

\subsection{Three Quality Paradigms}\label{subsec:quality}
Quality has several facets that no single metric settles, so we measure preservation under three paradigms, chosen so that none duplicates another. 

\textbf{Distributional (MAUVE).} We compute MAUVE~\cite{pillutla2021mauve} between $\mathcal{X}^+_c$ and $\mathcal{X}^-_c$. MAUVE captures shift in the whole output distribution but not meaning changes on individual prompts. 

\textbf{Paired-semantic (BERTScore).} We compute BERTScore F1~\cite{zhang2019bertscore} between each watermarked generation and the same prompt's unwatermarked completion, rescaled per language. BERTScore captures whether each prompt's meaning is kept but not a shift of the whole output distribution.

\textbf{Reference perplexity.} Under a held-out reference LM $R$ (XGLM-7.5B, with mGPT-1.3B as a cross-check), we compare the mean log-perplexity $\log \mathrm{PPL}_R(\cdot)$ of the watermarked and unwatermarked sets and report
\[
\mathrm{PPL\text{-}preservation}_c \;=\; \exp\!\left(-\bigl|\log \mathrm{PPL}_R(\mathcal{X}^+_c) - \log \mathrm{PPL}_R(\mathcal{X}^-_c)\bigr|\right) \;\in\; (0, 1].
\]
PPL captures how fluent the text stays under an external reference but inherits that reference's biases. The three paradigms can disagree on which schemes are least distorting (\S\ref{sec:findings}), so we report all three rather than pick one.

\subsection{Generalized-Entropy Decomposition}\label{subsec:ge}
We measure the disparity of a per-language vector $y$ of detection or quality scores by the generalized entropy index $\mathrm{GE}_2(y)$, half the squared coefficient of variation. Grouping languages by typological family splits $\mathrm{GE}_2$ exactly into within- and between-family parts~\cite{shorrocks1980class}. The between-family share shows whether gaps follow typology (script, morphology, tokenizer interactions) or fall language by language. Because six families contain a single language, a random relabeling already puts this share near 70\%, so we read it against the exact null of all 4{,}620 relabelings that keep the family sizes fixed (Appendix~\ref{app:sensitivity}). We register the typological partition of Table~\ref{tab:languages} before analysis and report a script-based partition as a robustness check.

\subsection{Disparity Measurements (M1-M4)}\label{subsec:m1m4} 
We apply four measurements to the per-language $\mathrm{TPR}_c(\alpha)$ and, under each quality paradigm, to per-language preservation (\S\ref{subsec:quality}). \textbf{M1} is the per-language vector. \textbf{M2} is its mean and minimum across languages. \textbf{M3} is the four-fifths disparate-impact count, the number of languages below $0.8\!\times\!\max$~\citep{feldman2015certifying}. \textbf{M4} is $\mathrm{GE}_2$ with its within/between-family split and shape-matched null (\S\ref{subsec:ge}).

\section{Findings}\label{sec:findings}

\subsection{Detection}
\label{sec:findings_detection}

Table~\ref{tab:detection_summary} summarizes detection across both regimes. In the base-FLORES panel, four of six schemes (KGW, SynthID, EXPEdit, DIP) reach mean TPR above 0.93 in every generator cell at $\tau_g$, with no per-language minimum below 0.798. The failures in this regime belong to Unigram and XSIR. The instruct-Aya panel reverses this picture, so we discuss the two regimes separately.

\begin{table*}[htbp]
\centering
\tiny
\setlength{\tabcolsep}{4pt}
\renewcommand{\arraystretch}{0.90}
\begin{tabular}{l l c c c c c c c c c}
\toprule
& & \multicolumn{2}{c}{AUC} & \multicolumn{5}{c}{TPR at $\tau_g$ (global)} & \multicolumn{2}{c}{TPR at $\tau_l$ (per-lang)} \\
\cmidrule(lr){3-4} \cmidrule(lr){5-9} \cmidrule(lr){10-11}
Scheme & Gen. & Mean & Min & Mean TPR & Min TPR & DI$_{<.8}$ & GE$_2$ & Btw.\,\% & Mean TPR & $\Delta$ \\
\midrule
\multicolumn{11}{l}{\textit{Base regime (FLORES+ continuation prompts)}} \\
\midrule
KGW     & Mistral & 0.999 & 0.997 & 0.992\,[0.987,\,0.996] & 0.978 & 0  & $<\!0.001$ & 83.6  & 0.991 & $+0.001$ \\
        & Gemma   & 0.997 & 0.991 & 0.969\,[0.959,\,0.979] & 0.934 & 0  & $<\!0.001$ & 72.4  & 0.974 & $-0.005$ \\
        & Qwen    & 0.996 & 0.987 & 0.965\,[0.939,\,0.986] & 0.880 & 0  & $<\!0.001$ & 85.1  & 0.966 & $-0.001$ \\
\cmidrule(lr){1-11}
Unigram & Mistral & 0.996 & 0.981 & 0.868\,[0.802,\,0.927] & 0.638 & 3  & 0.008      & 61.3  & 0.971 & $-0.103$ \\
        & Gemma   & 0.986 & 0.910 & 0.000\,[0.000,\,0.000] & 0.000 & 11 & ---        & ---   & 0.799 & $-0.799$ \\
        & Qwen    & 0.994 & 0.972 & 0.898\,[0.845,\,0.942] & 0.722 & 2  & 0.004      & 93.8  & 0.949 & $-0.051$ \\
\cmidrule(lr){1-11}
SynthID & Mistral & 1.000 & 1.000 & 1.000\,[0.999,\,1.000] & 0.998 & 0  & $<\!0.001$ & 100.0 & 1.000 & $-0.000$ \\
        & Gemma   & 1.000 & 1.000 & 1.000\,[0.999,\,1.000] & 0.998 & 0  & $<\!0.001$ & 100.0 & 1.000 & $+0.000$ \\
        & Qwen    & 1.000 & 0.999 & 0.997\,[0.995,\,0.999] & 0.990 & 0  & $<\!0.001$ & 97.5  & 0.998 & $-0.000$ \\
\cmidrule(lr){1-11}
EXPEdit & Mistral & 1.000 & 0.998 & 0.999\,[0.999,\,1.000] & 0.996 & 0  & $<\!0.001$ & 88.0  & 1.000 & $-0.000$ \\
        & Gemma   & 1.000 & 0.998 & 1.000\,[0.999,\,1.000] & 0.998 & 0  & $<\!0.001$ & 100.0 & 1.000 & $+0.000$ \\
        & Qwen    & 0.998 & 0.993 & 0.991\,[0.982,\,0.997] & 0.946 & 0  & $<\!0.001$ & 99.2  & 0.992 & $-0.001$ \\
\cmidrule(lr){1-11}
DIP     & Mistral & 0.997 & 0.989 & 0.975\,[0.966,\,0.984] & 0.938 & 0  & $<\!0.001$ & 92.6  & 0.975 & $+0.000$ \\
        & Gemma   & 0.992 & 0.965 & 0.963\,[0.947,\,0.977] & 0.914 & 0  & $<\!0.001$ & 99.5  & 0.965 & $-0.002$ \\
        & Qwen    & 0.992 & 0.977 & 0.931\,[0.891,\,0.968] & 0.798 & 0  & 0.003      & 94.7  & 0.929 & $+0.002$ \\
\cmidrule(lr){1-11}
XSIR    & Mistral & 0.981 & 0.962 & 0.784\,[0.743,\,0.821] & 0.672 & 2  & 0.004      & 91.5  & 0.833 & $-0.049$ \\
        & Gemma   & 0.984 & 0.977 & 0.766\,[0.712,\,0.817] & 0.588 & 4  & 0.006      & 96.0  & 0.829 & $-0.063$ \\
        & Qwen    & 0.978 & 0.930 & 0.215\,[0.118,\,0.362] & 0.050 & 10 & 0.493      & 98.9  & 0.794 & $-0.579$ \\
\midrule
\multicolumn{11}{l}{\textit{Instruct regime (Aya native-speaker prompts)}} \\
\midrule
KGW     & Mistral & 0.953 & 0.909 & 0.626\,[0.540,\,0.707] & 0.386 & 7  & 0.026      & 95.3  & 0.632 & $-0.005$ \\
        & Gemma   & 0.900 & 0.848 & 0.389\,[0.325,\,0.463] & 0.222 & 10 & 0.053      & 82.4  & 0.411 & $-0.022$ \\
        & Qwen    & 0.966 & 0.937 & 0.689\,[0.628,\,0.753] & 0.510 & 8  & 0.011      & 78.6  & 0.723 & $-0.034$ \\
\cmidrule(lr){1-11}
Unigram & Mistral & 0.909 & 0.865 & 0.311\,[0.171,\,0.453] & 0.022 & 6  & 0.297      & 99.8  & 0.450 & $-0.139$ \\
        & Gemma   & 0.851 & 0.790 & 0.037\,[0.023,\,0.052] & 0.010 & 10 & 0.226      & 81.2  & 0.151 & $-0.113$ \\
        & Qwen    & 0.943 & 0.880 & 0.546\,[0.455,\,0.621] & 0.210 & 4  & 0.032      & 91.7  & 0.636 & $-0.090$ \\
\cmidrule(lr){1-11}
SynthID & Mistral & 0.991 & 0.982 & 0.905\,[0.865,\,0.940] & 0.782 & 1  & 0.003      & 77.7  & 0.903 & $+0.002$ \\
        & Gemma   & 0.946 & 0.902 & 0.602\,[0.530,\,0.675] & 0.422 & 10 & 0.020      & 79.7  & 0.597 & $+0.005$ \\
        & Qwen    & 0.993 & 0.977 & 0.941\,[0.906,\,0.971] & 0.792 & 1  & 0.002      & 91.9  & 0.942 & $-0.002$ \\
\cmidrule(lr){1-11}
EXPEdit & Mistral & 0.918 & 0.782 & 0.577\,[0.454,\,0.698] & 0.238 & 6  & 0.067      & 80.0  & 0.591 & $-0.014$ \\
        & Gemma   & 0.621 & 0.462 & 0.225\,[0.179,\,0.275] & 0.084 & 10 & 0.071      & 84.2  & 0.237 & $-0.012$ \\
        & Qwen    & 0.839 & 0.683 & 0.496\,[0.379,\,0.620] & 0.174 & 9  & 0.080      & 84.4  & 0.508 & $-0.012$ \\
\cmidrule(lr){1-11}
DIP     & Mistral & 0.913 & 0.865 & 0.455\,[0.364,\,0.546] & 0.264 & 9  & 0.058      & 99.5  & 0.477 & $-0.022$ \\
        & Gemma   & 0.823 & 0.770 & 0.196\,[0.154,\,0.247] & 0.110 & 10 & 0.084      & 91.6  & 0.205 & $-0.010$ \\
        & Qwen    & 0.924 & 0.866 & 0.513\,[0.434,\,0.593] & 0.296 & 8  & 0.035      & 99.6  & 0.521 & $-0.007$ \\
\cmidrule(lr){1-11}
XSIR    & Mistral & 0.864 & 0.770 & 0.266\,[0.183,\,0.369] & 0.088 & 10 & 0.185      & 91.8  & 0.300 & $-0.035$ \\
        & Gemma   & 0.775 & 0.699 & 0.081\,[0.055,\,0.113] & 0.016 & 10 & 0.175      & 76.5  & 0.125 & $-0.043$ \\
        & Qwen    & 0.854 & 0.748 & 0.043\,[0.003,\,0.119] & 0.000 & 10 & 3.787      & 99.9  & 0.286 & $-0.243$ \\
\bottomrule
\end{tabular}
\caption{Detection fairness by scheme, generator, and regime, subset \texttt{all}, $\alpha = 0.01$. Mean TPR shown with 95\% bootstrap CIs ($B = 1000$ resamples of the 11-language vector). Min TPR is the realized minimum. DI$_{<.8}$ counts languages with $\mathrm{TPR}_l < 0.8 \cdot \max_l \mathrm{TPR}_l$. GE$_2$ uses the typological partition (Germanic, Romance, Indic, Semitic, Sinitic, Japonic, Turkic, Austroasiatic); Btw.\,\% is its between-family share (chance level ${\approx}70\%$, \S\ref{subsec:ge}). $\Delta = \mathrm{mean\,TPR}(\tau_g) - \mathrm{mean\,TPR}(\tau_l)$. Values below $10^{-3}$ reported as $<\!0.001$; entries marked --- are undefined when the per-language mean is zero.}
\label{tab:detection_summary}
\end{table*}

\textbf{Calibration versus detection failure.} Two base-regime cells that look like detection failures are calibration failures. Under single-threshold reporting, Unigram-Gemma shows mean TPR 0.000 with all eleven languages at zero, and XSIR-Qwen shows mean TPR 0.215 with ten of eleven languages flagged for disparate impact. Yet Unigram-Gemma's mean AUC is 0.986 with minimum 0.910, and XSIR-Qwen's mean AUC is 0.978, so the signal is present in both cells and the shared threshold sits too high for most watermarked text to clear. Calibrating a separate threshold for each language recovers mean TPR to 0.799 and 0.794, meaning one shared detector is the wrong deployment choice and not where the watermark failed. Without the companion of \S\ref{subsec:auc}, both cells would have been blamed on the watermark rather than on the threshold.

\textbf{Instruction tuning collapses detection.} Instruction tuning collapses detection at the shared threshold but not the watermark signal. Mean TPR falls below 0.7 in 16 of 18 instruct cells and DI$_{<.8} \geq 8$ in 12 of 18, with non-SynthID cells dropping roughly 0.17 to 0.78 in mean TPR from their base values. SynthID is the only partial exception. SynthID-Mistral and SynthID-Qwen retain mean TPR above 0.9, and SynthID-Gemma at 0.602 is the strongest Gemma instruct cell across all six schemes, though still well below its base value of 1.000. AUC nonetheless stays above the chance level of 0.5 in 197 of 198 instruct cell-language entries, the sole exception being EXPEdit-Gemma on Japanese at 0.462. The collapse is mostly a calibration problem. Instruction-tuned generation under native-speaker prompts has higher entropy, which pushes the two score distributions closer together without removing their separation. The base-versus-instruct axis is itself a fairness disparity, comparable to the cross-language gaps motivating the framework, and it holds across five of six scheme families.

\textbf{Detection disparity is structural to typological family.} Detection gaps follow language families more than chance alone would produce. For cells with GE$_2$ above the $10^{-3}$ floor, the between-family share averages 89.4\% (base, 6 cells) and 88.1\% (instruct, 18 cells). Because six families are singletons, a random same-shape relabeling yields about 70\% (95th percentile 83-84\%). Against that exact null both means are significant ($p = .0022$ and $.011$, Holm-adjusted $.0087$ and $.033$; Appendix~\ref{app:sensitivity}). The family partition captures more between-group variance than the script partition in every cell with non-trivial disparity (Table~\ref{tab:ge_partition_robustness}). XSIR-Qwen instruct, with GE$_2 = 3.787$, is an order of magnitude beyond any other entry. A leave-one-out check attributes it to Hindi alone (TPR 0.42 against $\leq 0.02$ elsewhere), and since Hindi is a singleton in both partitions, its 99.9\% between-family share is a single-language outlier, not a family pattern.

\textbf{On-target robustness.} The headline numbers include off-target generations by design (\S\ref{sec:grid}), so we recomputed the decomposition on the on-target subset with thresholds recalibrated on each cell's on-target pool (Table~\ref{tab:detection_on_target}). Every cell keeps all eleven languages, and the median between-family share moves by at most 2.5 points. Detection moves upward (base 94.2\% to 96.7\%, instruct 88.0\% to 90.0\%). MAUVE is essentially unchanged (63.4\% to 63.8\% and 90.0\% to 90.4\%). The detection result against the shape-matched null holds too ($p = .0035$ and $.0067$; Holm-adjusted at most $.014$ and $.027$). Per-cell values appear in Appendix~\ref{app:qstratify}. Filtering also separates the two base-regime failure cells. In Unigram-Gemma, off-target text in the pool used to set the threshold lifted the global threshold to 14.14 against 4.53 on-target; TPR recovers from 0.000 to 0.987, the cell falls below the disparity floor, and the failure lies with off-target text, not the watermark. In XSIR-Qwen the gap persists on correctly targeted text (mean TPR 0.215 to 0.207, GE$_2$ 0.49 to 0.53), so it belongs to the watermark rather than the generator.

\begin{figure}[htbp]
  \centering
  \includegraphics[width=\linewidth]{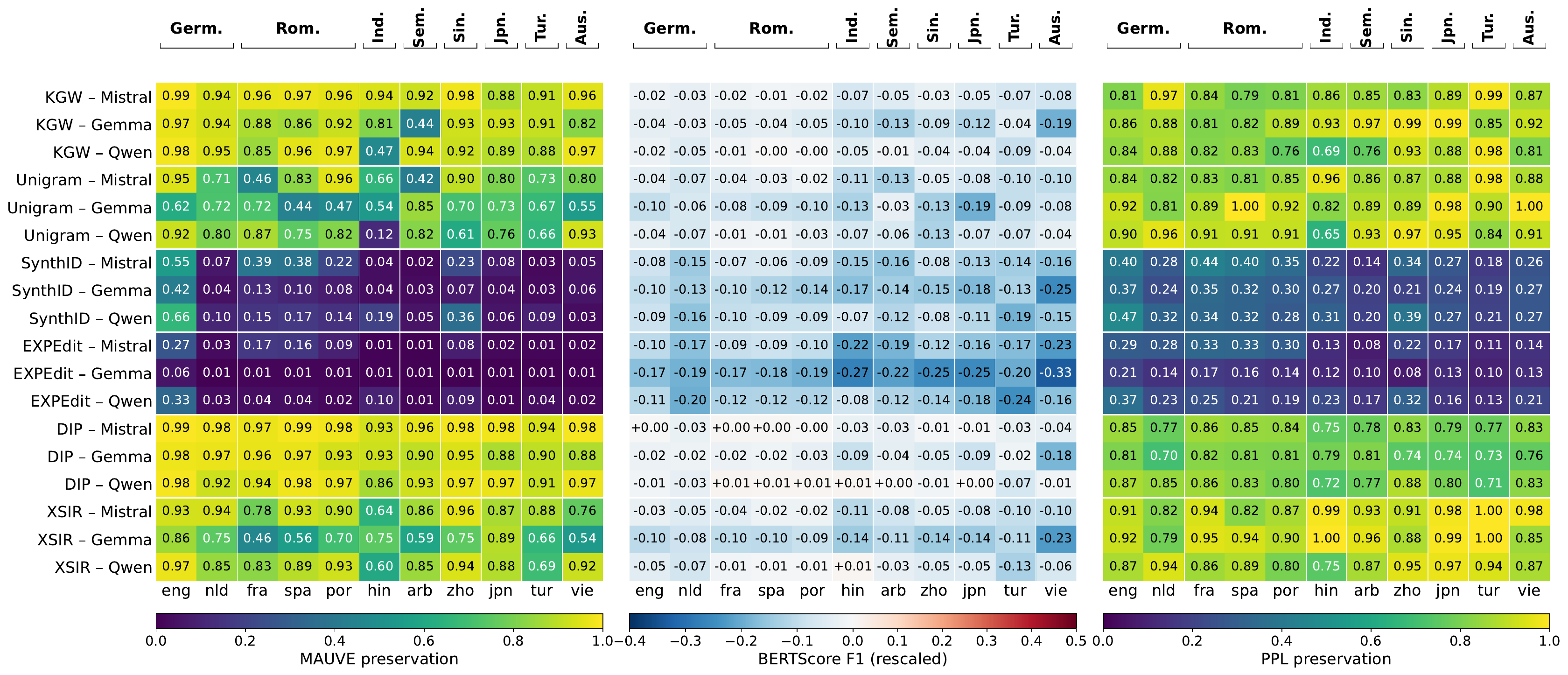}
  \includegraphics[width=\linewidth]{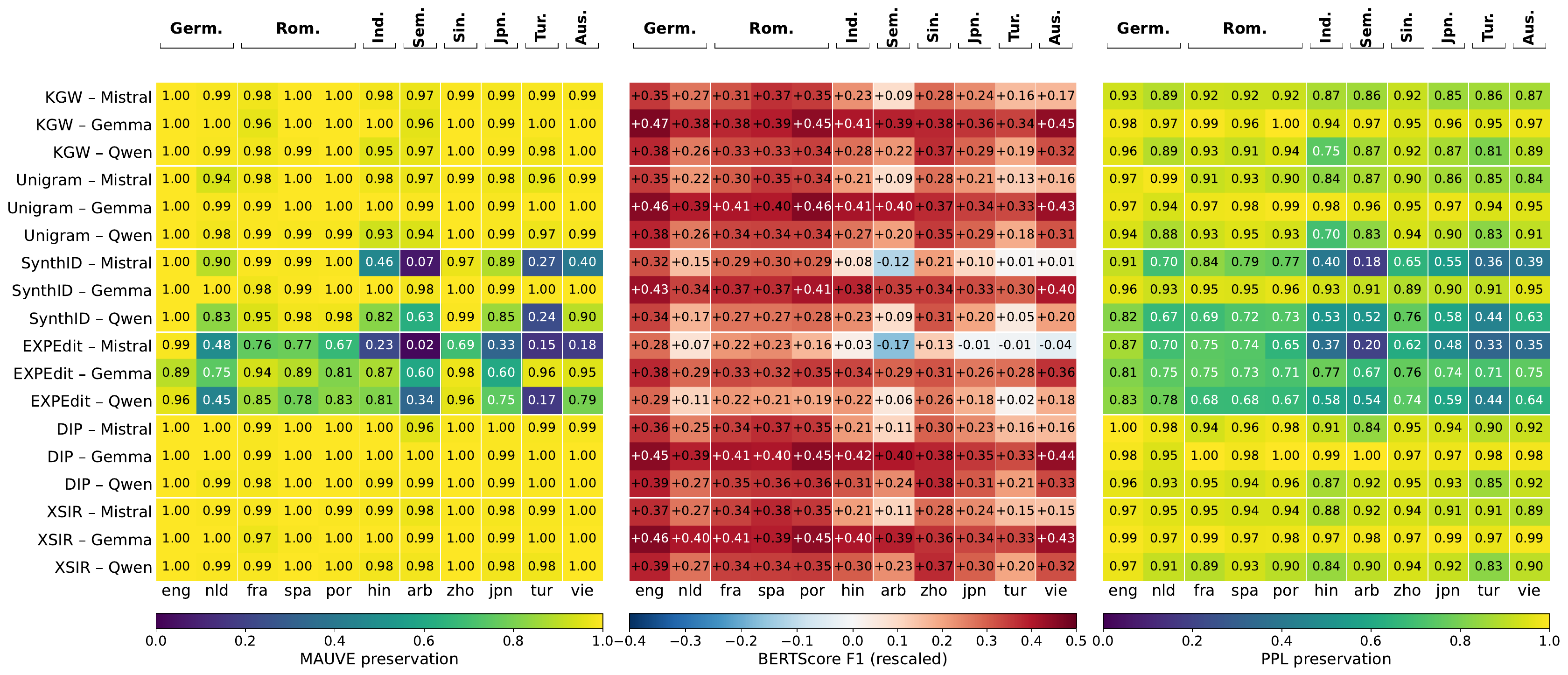}
  \caption{Per-language quality preservation under three paradigms, base-FLORES (top) and instruct-Aya (bottom); $n=500$ matched pairs per cell. Left: MAUVE on mean-pooled XLM-R-large embeddings ($1.0$ identical, $0.0$ fully separable). Center: BERTScore F1, rescaled against per-language FLORES non-pair baselines (negative: paired generations less similar than random non-pairs). Right: PPL preservation under XGLM-7.5B, $\exp(-|\log\mathrm{PPL}_R(\mathcal{X}^+) - \log\mathrm{PPL}_R(\mathcal{X}^-)|) \in (0,1]$. Rows: (scheme, generator) cells; columns: eleven languages grouped by typological family.}
  \label{fig:quality}
\end{figure}

\subsection{Quality}
\label{sec:quality}
\begin{figure}[htbp]
  \centering
  \includegraphics[width=0.95\linewidth]{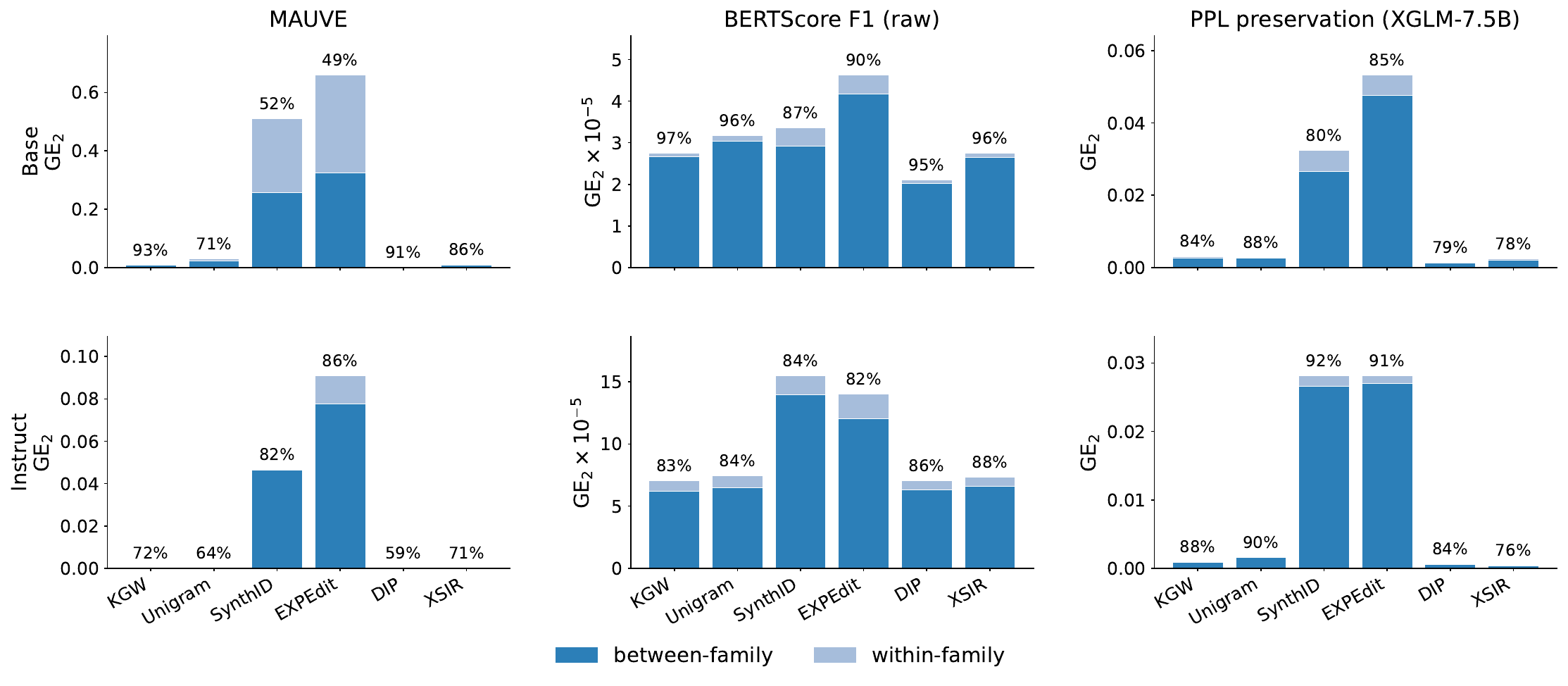}
\caption{Per-scheme disparity decomposition under three quality paradigms, base-FLORES (top) and instruct-Aya (bottom). Bar height is GE$_2$ over the per-language preservation vector, averaged across the three generators per scheme. Stacking shows the typological-partition decomposition into between-family (dark) and within-family (light) components; the percentage above is the between-family share (chance level ${\approx}70\%$, \S\ref{subsec:ge}). Under BERTScore, absolute disparity is orders of magnitude smaller than under MAUVE or PPL. Y-axes differ across panels.}
  \label{fig:ge2-decomposition}
\end{figure}

Figures~\ref{fig:quality} and~\ref{fig:ge2-decomposition} show per-language preservation under the three paradigms and its split into within- and between-family parts. Per-cell aggregates appear in Appendix~\ref{app:quality-summary}.

\textbf{Distortion-free schemes are the most distorting empirically.} EXPEdit and SynthID, the two sampling-time distortion-free schemes, damage quality most under every paradigm in the base regime. EXPEdit-Gemma, for instance, collapses to MAUVE 0.01-0.06 across its row (Figure~\ref{fig:quality}, top panels; per-cell values in Table~\ref{tab:q_5_2_summary}). The distortion-free guarantee covers the next-token distribution on average, not any of the three measurements on finite samples. The four logit-bias schemes (KGW, Unigram, DIP, XSIR) all preserve quality substantially better.

\textbf{Instruct regime compression.} Instruction tuning pushes all three quality measurements upward without changing which schemes are worst. MAUVE reaches its ceiling on KGW, Unigram, DIP, and XSIR, and PPL preservation rises on the same four (Table~\ref{tab:q_5_2_summary}). BERTScore-rescaled values turn large and positive because Aya prompts make even unrelated outputs alike, which lowers the baseline they are rescaled against. The compression parallels the detection-side compression of \S\ref{sec:findings_detection}, and SynthID and EXPEdit remain the most distorting. Within schemes it tracks the detection collapse. Cells where the watermark detects weakly (e.g., SynthID-Gemma instruct, TPR 0.602) show near-perfect quality preservation, consistent with the watermark changing the generation distribution less.

\textbf{Quality disparity is large but only weakly family-structured.} Quality gaps are large but do not follow language family beyond chance. Figure~\ref{fig:ge2-decomposition} reports GE$_2$ over the per-language preservation vector, averaged over the three generators per scheme, stacked into between- and within-family parts. On MAUVE, SynthID and EXPEdit dominate cross-language disparity, with GE$_2$ peaking at 0.66 for EXPEdit base, and on PPL the peak is $\approx$0.053, an order of magnitude beyond any logit-bias scheme. They distort most on average and most unevenly across languages. Between-family shares are $\geq 70\%$ in 32 of 36 (scheme, paradigm, regime) cells, but 70\% is roughly the chance level (\S\ref{subsec:ge}). Against the exact null, MAUVE shares sit at chance in base (mean 69.3\% vs.\ null 71.6\%, $p = .57$). Under instruct they clear it only before correction (90.8\% vs.\ 68.9\%, $p = .039$, Holm $.078$), so we do not claim quality disparity is family-structured.

\textbf{Cross-paradigm rank agreement.} The paradigms often rank a scheme's languages differently and the disagreement depends on where the watermark acts (Table~\ref{tab:q_5_2_qrob}). Pooled across the grid, no paradigm pair clears $\rho = 0.6$: MAUVE-vs-BERTScore averages $+0.58$, MAUVE-vs-PPL $+0.38$, and BERTScore-vs-PPL $+0.37$. The two sampling-time distortion-free schemes (SynthID, EXPEdit) sit at $\rho \in [0.68, 0.81]$ across all three pairs with strictly positive cell-level minima, so the three paradigms agree on the language ordering. The three green-list schemes (KGW, Unigram, XSIR) sit at PPL-vs-embedding mean correlations of $\leq +0.20$ with cell-level minima between $-0.39$ and $-0.80$, marking cells where reference perplexity and the embedding-based paradigms order languages oppositely. DIP, distortion-free in design but logit-side in implementation, falls in between at $\rho \in [0.43, 0.50]$. Where the watermark intervenes at the logit level, PPL measures fluency under a held-out reference while MAUVE and BERTScore measure divergence from the unwatermarked set, and need not move together. With Hindi being the cleanest instance, KGW, Unigram, and XSIR all preserve Hindi PPL at $\geq 0.65$ under both XGLM and mGPT references (Appendix~\ref{app:ppl-mgpt}), while MAUVE flags outliers on the same cells (Unigram-Qwen 0.12, KGW-Qwen 0.47).

\begin{table}[htbp]
\centering
\scriptsize
\begin{tabular}{lccc}
\toprule
Scheme & MAUVE $\leftrightarrow$ BS & MAUVE $\leftrightarrow$ PPL & BS $\leftrightarrow$ PPL \\
\midrule
KGW            & +0.61 (+0.45) & +0.03 ($-$0.80) & +0.03 ($-$0.76) \\
Unigram        & +0.56 (+0.23) & +0.14 ($-$0.39) & +0.20 ($-$0.67) \\
SynthID        & +0.76 (+0.47) & +0.81 (+0.27)   & +0.77 (+0.45)   \\
EXPEdit        & +0.68 (+0.26) & +0.75 (+0.39)   & +0.70 (+0.46)   \\
DIP            & +0.46 ($-$0.03) & +0.43 (+0.05) & +0.50 ($-$0.05) \\
XSIR           & +0.42 ($-$0.09) & +0.10 ($-$0.71) & +0.00 ($-$0.76) \\
\midrule
\textit{All schemes} & +0.58 ($-$0.09) & +0.38 ($-$0.80) & +0.37 ($-$0.76) \\
\bottomrule
\end{tabular}
\caption{Cross-paradigm rank agreement: per-cell Spearman $\rho$ between per-language preservation vectors, mean across (model, regime) cells with per-cell minimum in parentheses (subset=\texttt{all}; $n=6$ cells per scheme, $n=36$ pooled). Pairs use BERTScore F1 (rescaled) and PPL preservation under XGLM-7.5B; BS-raw gives nearly identical rankings ($\rho_{\text{BS-raw} \leftrightarrow \text{BS-rescaled}} = +0.87$). Negative minima identify cells where two paradigms produce opposite per-language orderings.}
\label{tab:q_5_2_qrob}
\end{table}

\subsection{Joint Detection-Quality}
\label{sec:joint}
\begin{figure}[htbp]
\centering
\includegraphics[width=\textwidth]{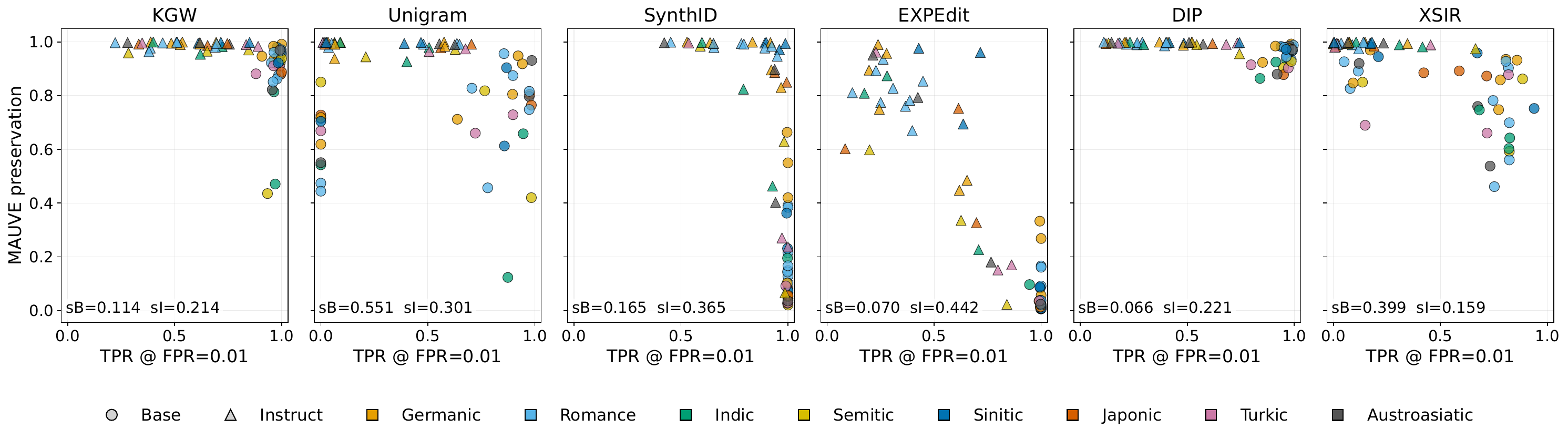}
\caption{Joint per-language detection-quality landscape, MAUVE paradigm. Markers: per-language $(\mathrm{TPR}\,@\,\tau_g, \mathrm{MAUVE})$ pairs at $\alpha{=}0.01$. Circles: base-FLORES. Triangles: instruct-Aya. Color: typological family. Annotations $s_B$, $s_I$: mean pairwise Euclidean distance within each regime. BERTScore and PPL-preservation companion panels in Appendix~\ref{app:joint}.}
\label{fig:pareto-headline}
\end{figure}

Detection (\S\ref{sec:findings_detection}) and quality (\S\ref{sec:quality}), read separately, do not show how they combine. A scheme balanced on each can still be unfair if the languages it detects well are not the ones it preserves well. Figure~\ref{fig:pareto-headline} plots each language's $(\mathrm{TPR}\,@\,\tau_g, \mathrm{MAUVE})$ pair per scheme and annotates each regime's spread $s$, the mean pairwise Euclidean distance over its 33 points (11 languages $\times$ 3 generators).

\textbf{Spread is regime-dependent and must be read with its panel.} DIP base ($s_B{=}0.066$) and EXPEdit base ($s_B{=}0.070$) have near-identical spread but sit in opposite corners. DIP is top right, detecting and preserving well, and EXPEdit bottom right, detecting everywhere and preserving nowhere. XSIR instruct ($s_I{=}0.159$) is tight too, at the top left under uniform detection collapse. For base, the spread is attributed to calibration failure. Unigram ($s_B{=}0.551$) and XSIR ($s_B{=}0.399$) scatter along the detection axis with quality intact, as in \S\ref{sec:findings_detection}, while the other four panels are tight ($s_B \le 0.165$). Under instruct, EXPEdit ($s_I{=}0.442$), SynthID ($s_I{=}0.365$), and Unigram ($s_I{=}0.301$) lead as detection collapse scatters some languages along the quality axis. The regime does not change the size but rather changes the shape of joint unfairness.

\textbf{Who is underserved.} No single language is worst everywhere, but on quality the floor is never English. The four measures of \S\ref{subsec:m1m4}, read in order, locate the worst-served languages and say whether they cluster by family. On quality, the worst-served language is Arabic, Hindi, Turkish, Vietnamese, or Japanese in at least four cells each across the grid (Table~\ref{tab:q_5_2_summary}; these are the columns to read in Figure~\ref{fig:quality}). On detection, the worst-served language changes with scheme and regime (Table~\ref{tab:bottom_quintile}), so no language can be declared safe in general, and fairness has to be audited for each deployed (scheme, generator, regime) configuration, which a deployer would choose.

\textbf{Why the gaps arise.} Detection gaps follow how well a model holds the prompted language, and quality gaps follow how many tokens a language needs. Appendix~\ref{app:mechanism} regresses per-language detection and quality on tokenizer fertility (tokens per sentence), unwatermarked surprisal, a Latin-script indicator, and language adherence. A watermark embeds its signal in the model's choice at each next token, so it needs available choice to steer, and detection accordingly follows adherence and entropy. Adherence is the dominant base-regime predictor for KGW, SynthID, and DIP ($R^2$ between 0.41 and 0.60). Fertility is dominant for EXPEdit (0.83), consistent with the known dependence of detection on entropy~\cite{kirchenbauer2023watermark, lee-etal-2024-wrote}. Quality instead follows fertility for the green-list schemes (base MAUVE $R^2$ up to 0.39, instruct PPL up to 0.56). A language cut into more pieces per sentence receives more watermark interventions per sentence, so distortion accumulates unevenly, in line with the uneven vocabulary allocation documented for subword tokenizers~\cite{rust-etal-2021-good, petrov2023language, ahia-etal-2023-languages}. The writing system itself is not the driver, as the Latin-script indicator and surprisal clear $p<0.05$ in 5 of 48 cells combined, where they are collinear with family.

\section{Discussion}\label{discussion}

\textbf{Limitations:} Our eleven languages span four scripts and eight typological families, but many of the world's languages remain outside this set. Six families are represented by a single language, so family-level conclusions rest on one sampled language each and may not generalize~\cite{ploeger-etal-2026-principled}. Our generators are 12B parameters and below; the Qwen2.5-72B pilot of Appendix~\ref{app:scale} covers three schemes in the base regime only. Our six schemes cover the dominant green-list and distortion-free design families but are not exhaustive of an actively growing literature.

\textbf{Societal impacts:} Watermarking is already deployed as a scalable mitigation for AI-generated text, much of which is multilingual. A detection gap that follows typological family rather than a single language means the same scheme gives speakers of some languages materially weaker protection than others. Quality gaps, whether family-structured or not, mean the cost of being watermarked is unequally borne. Because the detection gaps follow families, they are less likely to close by adding data for individual languages and instead implicate architectural choices that act above any one language, such as tokenizer construction or partition design. The tokenizer link agrees with prior evidence that subword tokenizers allocate vocabulary unevenly across languages~\cite{rust-etal-2021-good, petrov2023language, ahia-etal-2023-languages}. As \S\ref{sec:joint} and Appendix~\ref{app:mechanism} show, tokenizer fertility tracks the quality gaps and language adherence the detection gaps. Appendix~\ref{app:levers} ranks the candidate design axes and maps mitigations to who controls them: per-language calibration for deployers, fertility-aware green-list and partition construction for scheme designers, and balanced vocabularies and multilingual coverage for model developers.

\textbf{Deployment guidance:} In the base regime, DIP alone pairs near-ceiling detection with no language flagged for disparate impact on quality (Table~\ref{tab:q_5_2_summary}; Figure~\ref{fig:pareto-headline}, top-right). SynthID and EXPEdit detect essentially everywhere at the largest, most uneven quality cost, which suits settings where detection outweighs text fidelity. Under instruction tuning, only SynthID retains mean TPR above 0.9, and only on two of three generators, so the regime is itself a deployment variable.

\textbf{Model scale:} Appendix~\ref{app:mechanism} suggests that detection gaps should narrow with scale, because they track language adherence and larger models hold the prompted language more reliably. It also suggests that quality gaps should persist, because they track tokenizer fertility, which does not change with parameter count. A Qwen2.5-72B pilot of KGW, SynthID, and XSIR in the base regime (Appendix~\ref{app:scale}) confirms neither prediction. The worst-language TPR moves by a median of $-1.0$~pp, and XSIR remains the lowest-TPR scheme. Quality levels rise (median MAUVE $+0.052$) although the 72B model shares the 7B generator's tokenizer. With at most three cells per test, the between-family shares sit above chance without reaching significance. Scale behavior therefore remains a hypothesis, and the instruct regime is untested.

\section{Conclusion} In this work, we introduce the first systematic cross-lingual fairness audit of LLM text watermarking: a four-component evaluation framework comprising empirical FPR calibration, a threshold-independent AUC companion, three disjoint quality paradigms, and a generalized-entropy decomposition over a typological partition. We apply this framework to six schemes, three open-weight generators, and eleven languages across four scripts and eight typological families, under two generation regimes. The audit surfaces failure modes that single-language, single-paradigm evaluation cannot, including calibration failures misread as detection failures, distortion-free schemes that are empirically the most distorting, cross-paradigm rank inversion, and detection disparity dominated by between-family variation. Our work promotes further research toward fairness in LLM watermarking.

\section*{Acknowledgments}
This work was supported by the National Science Foundation (NSF) under grant numbers OAC-2112606, 2427505, and 2141622, by the National Library of Medicine of the National Institutes of Health (NIH) under award number R01LM014520, and by the U.S. Department of Agriculture (USDA) National Institute of Food and Agriculture under award number 2026-67023-46766.

\bibliographystyle{abbrv}
\bibliography{references}


\appendix

\section{Code Availability and Compute}\label{ai_usage}

\textbf{Code.} Code for the evaluation framework, generation pipeline, and analysis is available here: \url{https://github.com/ANCP2021/Cross-Lingual-Watermarking-Fairness-Audit}.

\textbf{Compute.} All generation and evaluation runs were conducted on a mix of NVIDIA H100 and A100 GPUs.

\section{Framework Details}\label{app:framework-details}
This section collects the formal definitions behind \S\ref{sec:framework}.

\textbf{Thresholds (\S\ref{subsec:tau}).} Each scheme $s$ defines a score $S_s : \mathcal{T} \to \mathbb{R}$ on texts, with watermarked text receiving systematically larger scores than unwatermarked. Each of the six schemes ships a hardcoded threshold targeting a theoretical FPR under an IID-token null that cross-lingual generation does not satisfy; \S\ref{sec:findings_detection} exhibits a scheme-language pair where the null tail saturates at the score cap so no watermarked text can clear the default. The per-language threshold $\tau^{\mathrm{lang}}_c(\alpha)$ is $\tau_c(\alpha)$ computed on $\mathcal{X}^-_c$ alone, and the global threshold $\tau^{\mathrm{global}}_{(s,m,r)}(\alpha)$ is computed on the pooled null $\bigcup_l \mathcal{X}^-_{(l,s,m,r)}$; tables and figures write these as $\tau_l$ and $\tau_g$. Cells with insufficient null samples are excluded from the disparity rollups, with the sample-size floors stated in \S\ref{sec:grid}.

\textbf{Companion measurement (\S\ref{subsec:auc}).} $\mathrm{AUC}_c$ is estimated by the Mann-Whitney $U$ statistic with average-rank tie handling.

\textbf{Quality paradigms (\S\ref{subsec:quality}).} MAUVE is computed on mean-pooled XLM-R-large embeddings. BERTScore F1 is rescaled with per-language baselines estimated from FLORES non-pairs. For reference perplexity, the held-out reference LM $R$ is XGLM-7.5B as primary and mGPT-1.3B as cross-reference (Appendix~\ref{app:ppl-mgpt}), and $\log \mathrm{PPL}_R(\mathcal{X})$ denotes the macro-averaged log-perplexity of the generations in $\mathcal{X}$ under $R$.

\textbf{Generalized entropy (\S\ref{subsec:ge}).} For a per-language vector $y = (y_1, \ldots, y_L)$, the generalized entropy index at $\alpha = 2$ is $\mathrm{GE}_2(y) = (2L)^{-1} \sum_i \! \left[ (y_i / \bar{y})^2 - 1 \right]$, which equals half the squared coefficient of variation. Under a partition of languages into typological families $\mathcal{F}$, $\mathrm{GE}_2$ admits the standard within/between decomposition $\mathrm{GE}_2(y) = \mathrm{GE}_2^{\mathrm{within}}(y) + \mathrm{GE}_2^{\mathrm{between}}(y)$~\cite{shorrocks1980class}.

\section{Generation Preprocessing}\label{preprocessing-app}

\subsection{Language Adherence}
\label{sec:p1-adherence}

Table~\ref{tab:p1-adherence} reports per-cell on-target rates from GlotLID-v3 for unwatermarked generations (NWM\%) and the mean shift across the six watermarked schemes ($\bar{\Delta}$, percentage points). Baseline adherence varies systematically: generator-mean rates run from 80.9\% (Mistral instruct) to 96.5\% (Mistral base), with cell-level minima at Qwen-Turkish base (57.2\%) and Mistral-Hindi instruct (52.2\%). Instruction tuning does not uniformly raise adherence; Mistral and Qwen lose ground on several non-Latin-script cells while Gemma remains roughly stable.

Watermark-induced shifts are modest in base (2 of 33 cells exceed $|\bar{\Delta}| = 5$~pp) and larger in instruct (12 of 33), concentrated on Mistral and Qwen non-Latin cells while Gemma instruct stays close to its baseline ($\bar{\Delta} = -0.4$~pp on average). Three instruct cells lose more than 13 points: Mistral-Arabic ($-15.6$), Qwen-Arabic ($-15.0$), and Qwen-Turkish ($-13.5$). These are the cells where the on-target subset thins toward the $n = 200$ sample-size floor for $\alpha = 0.01$ calibration, and they produce the largest subset-\texttt{all}-vs-on-target gaps in the detection robustness analysis.

\begin{table*}[htbp]
\centering
\tiny
\setlength{\tabcolsep}{4pt}
\begin{tabular}{l rr rr rr c rr rr rr}
\toprule
 & \multicolumn{6}{c}{\textit{Base regime (FLORES+ continuation prompts)}} & & \multicolumn{6}{c}{\textit{Instruct regime (Aya native-speaker prompts)}} \\
\cmidrule(lr){2-7} \cmidrule(lr){9-14}
 & \multicolumn{2}{c}{Mistral} & \multicolumn{2}{c}{Gemma} & \multicolumn{2}{c}{Qwen} & & \multicolumn{2}{c}{Mistral} & \multicolumn{2}{c}{Gemma} & \multicolumn{2}{c}{Qwen} \\
\cmidrule(lr){2-3}\cmidrule(lr){4-5}\cmidrule(lr){6-7}\cmidrule(lr){9-10}\cmidrule(lr){11-12}\cmidrule(lr){13-14}
Lang & NWM\% & $\bar{\Delta}$ & NWM\% & $\bar{\Delta}$ & NWM\% & $\bar{\Delta}$ & & NWM\% & $\bar{\Delta}$ & NWM\% & $\bar{\Delta}$ & NWM\% & $\bar{\Delta}$ \\
\midrule
eng & 99.6 & $-0.3$ & 98.8 & $-2.0$ & 98.4 & $-0.6$ & & 92.6 & $-0.9$ & 80.6 & $-1.7$ & 93.2 & $-2.0$ \\
fra & 99.2 & $-1.1$ & 97.6 & $-4.0$ & 89.2 & $-0.2$ & & 91.4 & $+0.5$ & 86.2 & $-1.7$ & 94.8 & $-4.9$ \\
spa & 100.0 & $-1.4$ & 96.8 & $-2.6$ & 91.2 & $+1.1$ & & 95.2 & $-1.3$ & 62.6 & $+2.7$ & 92.0 & $-5.0$ \\
por & 99.2 & $-0.5$ & 96.8 & $-1.0$ & 98.2 & $-0.8$ & & 96.2 & $-2.0$ & 70.6 & $+3.0$ & 96.4 & $-4.3$ \\
nld & 96.2 & $-7.0$ & 95.8 & $+1.6$ & 80.8 & $-4.6$ & & 73.2 & $-3.1$ & 89.2 & $-0.4$ & 90.4 & $-5.8$ \\
zho & 97.6 & $-1.8$ & 92.8 & $-1.2$ & 96.4 & $-4.2$ & & 89.0 & $-7.9$ & 95.6 & $-0.4$ & 98.4 & $-0.6$ \\
jpn & 96.8 & $-3.3$ & 95.6 & $-1.6$ & 99.8 & $-3.3$ & & 71.0 & $-10.7$ & 94.0 & $-1.9$ & 98.4 & $-7.4$ \\
arb & 89.2 & $-3.0$ & 87.6 & $-3.8$ & 87.8 & $+2.9$ & & 67.6 & $-15.6$ & 84.2 & $-1.2$ & 81.0 & $-15.0$ \\
hin & 96.2 & $-9.1$ & 88.2 & $-2.3$ & 98.8 & $-3.9$ & & 52.2 & $-6.5$ & 83.8 & $-1.4$ & 83.2 & $+0.1$ \\
tur & 89.0 & $+1.3$ & 96.8 & $-4.3$ & 57.2 & $-1.1$ & & 81.8 & $-8.1$ & 92.8 & $-0.6$ & 95.6 & $-13.5$ \\
vie & 98.6 & $-4.7$ & 77.0 & $+4.7$ & 99.0 & $-1.0$ & & 79.2 & $-7.3$ & 98.8 & $-0.6$ & 95.2 & $-6.4$ \\
\midrule
mean & 96.5 & $-2.8$ & 93.1 & $-1.5$ & 90.6 & $-1.4$ & & 80.9 & $-5.7$ & 85.3 & $-0.4$ & 92.6 & $-5.9$ \\
\bottomrule
\end{tabular}
\caption{Per-cell language adherence on the eleven evaluation languages, base-FLORES (left) and instruct-Aya (right). NWM\% is the GlotLID-v3 on-target rate for no watermark generations; $\bar{\Delta}$ is the mean shift in percentage points across the six watermarked schemes. The \textit{mean} rows are unweighted means across the eleven languages.}
\label{tab:p1-adherence}
\end{table*}

\subsection{Generation Length Statistics}
\label{sec:p1-lengths}
Generations sit close to the maximum-token cap in the base regime where language-mean lengths are $196.7 \pm 16.2$ (Mistral), $196.6 \pm 16.8$ (Gemma), and $187.3 \pm 26.6$ (Qwen) tokens and shift to shorter, more dispersed distributions under instruction tuning ($167.7 \pm 37.8$, $174.6 \pm 39.4$, $180.2 \pm 32.0$ respectively). Per-cell variation is largest on Mistral instruct, where mean lengths range from 152.2 tokens (Portuguese) to 183.2 (English).

\begin{table}[htbp]
\centering
\small
\begin{tabular}{llrrrrrr}
\toprule
Regime & Subset & $n$ kept & $n$ skipped & Mean & Median & Min & Max \\
\midrule
Base & on\_target\_only & 231 & 0 & 1.003 & 1.000 & 0.996 & 1.081 \\
Base & off\_target\_only & 8 & 223 & 0.929 & 0.908 & 0.859 & 1.023 \\
Instruct & on\_target\_only & 231 & 0 & 1.000 & 1.000 & 0.964 & 1.057 \\
Instruct & off\_target\_only & 26 & 205 & 1.025 & 1.017 & 0.947 & 1.119 \\
\bottomrule
\end{tabular}
\caption{Length-distribution similarity between LID subsets and the all-subset baseline. Each row is a (regime, subset) pair; ratio$=$subset-mean length/all-subset mean length, computed per (regime, model, language, watermark) cell and summarized across cells. Cells require $n \geq 100$ in both subsets to enter the summary; \textit{$n$ skipped} counts cells that fail this floor (typically off\_target on high-adherence cells).}
\label{tab:p1-length-subset}
\end{table}

Table~\ref{tab:p1-length-subset} reports the ratio of subset-conditioned mean length to all-subset mean length, computed per (regime, model, language, watermark) cell with $n \geq 100$ in both subsets. On-target subset lengths are within 0.5\% of all-subset means in aggregate (median ratio $1.000$ in both regimes; per-cell range $0.964$-$1.081$), so the on-target stratification used downstream is not confounded by systematic length differences. The off-target subset clears the $n \geq 100$ floor on only 8 of 231 base cells and 26 of 231 instruct cells. This is a consequence of high baseline adherence outside the cells flagged in \S\ref{sec:p1-adherence} but where measurable, its mean lengths track all-subset means within $\sim7$\%.

\subsection{Categorization Robustness}\label{sec:p1-cat-robustness}
The on-target categorization depends on two thresholds: a GlotLID per-sentence confidence floor (\texttt{conf\_floor}) and a per-text on-target sentence fraction (\texttt{on\_target\_ratio}). Table~\ref{tab:p1-threshold-sweep} sweeps these on a $3 \times 3$ grid and reports Spearman rank-correlation against the default-config per-cell adherence vector. Rank-correlation stays $\geq0.86$ across all eight alternatives, and the mean on-target rate across cells moves by at most 7.3~pp ($82.5\%$ at the strictest config to $92.0\%$ at the most permissive). The cell-level adherence patterns that drive \S\ref{sec:p1-adherence} are not knife-edge sensitive to the threshold choice.

\begin{table}[htbp]
\centering
\scriptsize
\begin{tabular}{rrrrrr}
\toprule
\texttt{conf\_floor} & \texttt{on\_target\_ratio} & $\rho$ vs default & Mean on-target \% & Mean $|\Delta|$ (pp) & Max $|\Delta|$ (pp) \\
\midrule
0.3 & 0.7 & 0.9347 & 92.0 & 2.19 & 13.60 \\
0.3 & 0.8 & 0.9468 & 90.7 & 0.91 & 13.00 \\
0.3 & 0.9 & 0.8627 & 86.3 & 4.87 & 31.00 \\
0.5 & 0.7 & 0.9910 & 91.1 & 1.26 & 11.80 \\
\textbf{0.5}$^\dagger$ & \textbf{0.8} & \textbf{1.0000} & \textbf{89.8} & \textbf{0.00} & \textbf{0.00} \\
0.5 & 0.9 & 0.9317 & 85.4 & 4.43 & 31.20 \\
0.7 & 0.7 & 0.9486 & 87.9 & 3.76 & 24.60 \\
0.7 & 0.8 & 0.9627 & 86.8 & 3.04 & 24.80 \\
0.7 & 0.9 & 0.9385 & 82.5 & 7.36 & 31.80 \\
\bottomrule
\end{tabular}
\caption{LID categorization robustness: per-cell adherence under a $3 \times 3$ sweep of the GlotLID per-sentence confidence floor (\texttt{conf\_floor}) and the per-text on-target sentence fraction (\texttt{on\_target\_ratio}). $\rho$ is the Spearman rank-correlation between the per-cell on-target-rate vector under each configuration and the default ($\dagger$). \textit{Mean on-target \%} is the unweighted mean across the 462 cells of the evaluation grid. \textit{Mean/Max $|\Delta|$} are the mean and maximum absolute per-cell shift in on-target rate (pp) relative to default.}
\label{tab:p1-threshold-sweep}
\end{table}

\section{Detection: Per-Language Detail and Robustness}
\label{app:detection}
\begin{figure}[htbp]
\centering
\begin{subfigure}[htbp]{0.49\linewidth}
\centering
\includegraphics[width=\linewidth]{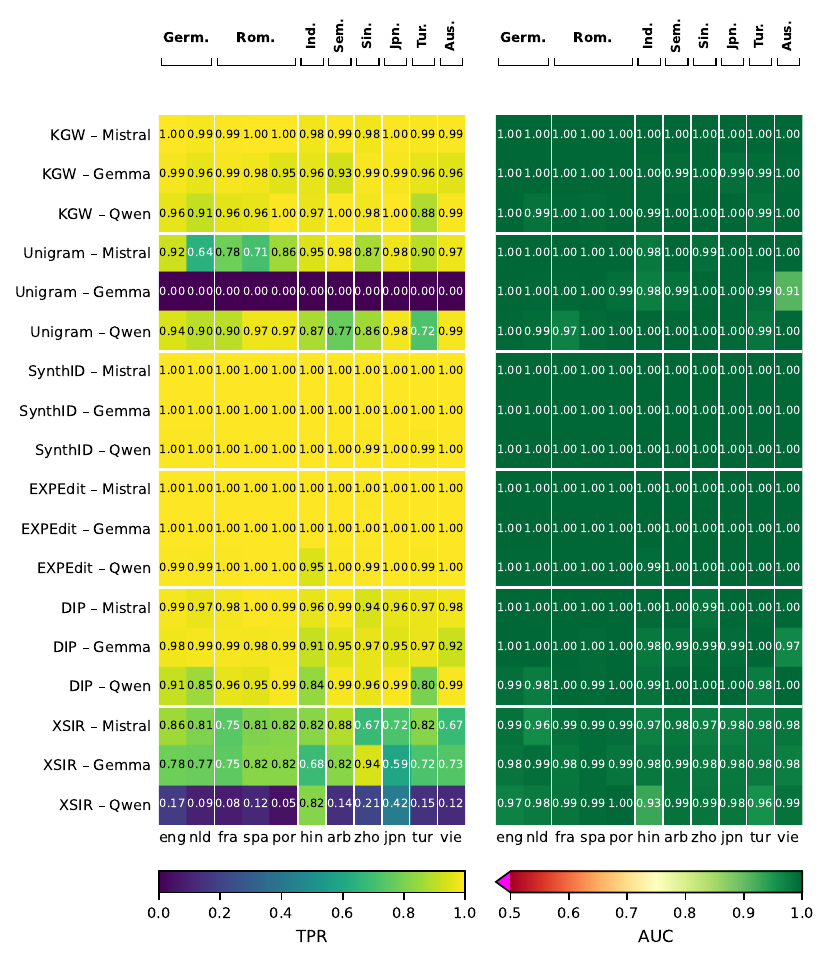}
\caption{Base-FLORES regime.}
\label{fig:detection_heatmap_base}
\end{subfigure}
\hfill
\begin{subfigure}[htbp]{0.49\linewidth}
\centering
\includegraphics[width=\linewidth]{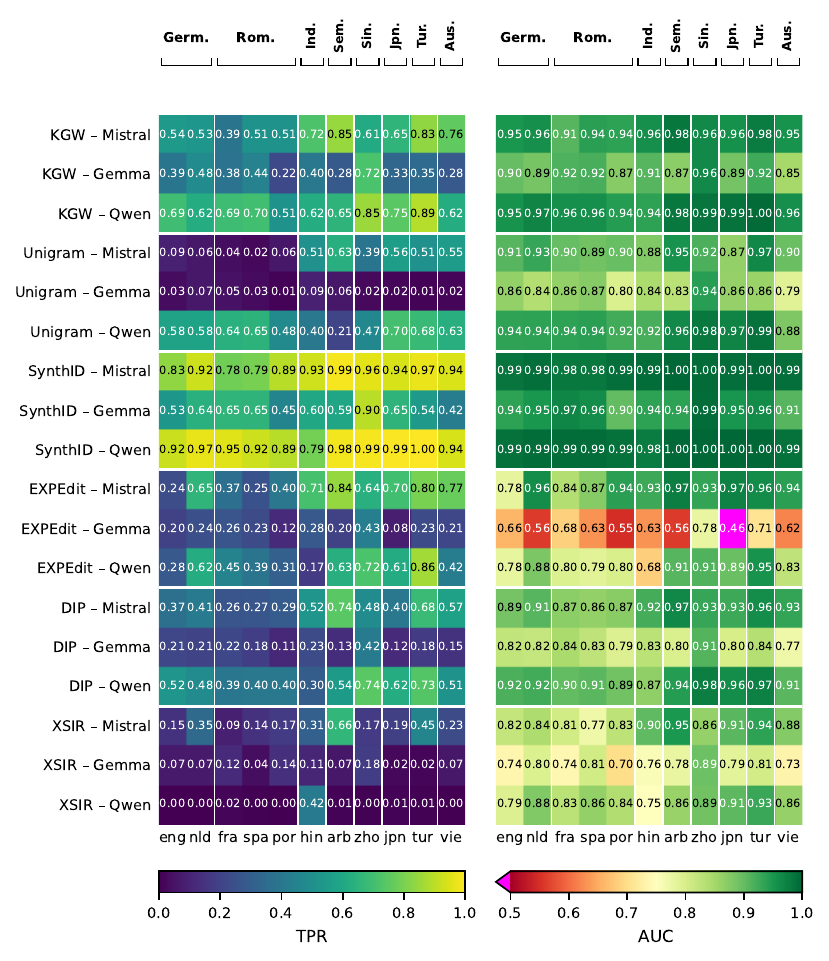}
\caption{Instruct-Aya regime.}
\label{fig:detection_heatmap_instruct}
\end{subfigure}
\caption{Per-language detection diagnostics, subset \texttt{all}, $\alpha = 0.01$. Within each panel, left: TPR at the global empirical-FPR threshold $\tau_g$; right: threshold-free AUC. Rows are (scheme, generator) cells; columns are the eleven evaluation languages, grouped by typological family.}
\label{fig:detection_heatmap}
\end{figure}

This appendix supplies the detection-side material that the main text's headline summary (Table~\ref{tab:detection_summary}) and \S\ref{sec:findings_detection} collapse or defer. It is organized around four questions, each answered by one or two tables. First, what does the per-language vector behind each headline cell actually look like (Figures~\ref{fig:detection_heatmap_base} and~\ref{fig:detection_heatmap_instruct}, Table~\ref{tab:detection_calibration_gap})? Second, do the calibration-failure cells flagged in \S\ref{sec:findings_detection} survive alternative operating choices on the FPR target and the language-adherence subset (Tables~\ref{tab:detection_alpha_05}, \ref{tab:detection_on_target})? Third, do the per-cell disparity statistics depend on whether Min TPR is read as a single-language outlier or a small low-performing cluster (Table~\ref{tab:bottom_quintile})? Fourth, does the family-structure claim of \S\ref{sec:findings_detection} survive swapping the inequality measure (GE$_2 \to$ GE$_0$) or the typological partition (family $\to$ script) used in the decomposition (Tables~\ref{tab:ge_0_robustness}, \ref{tab:ge_partition_robustness}; \S\ref{app:sensitivity})? Each table's caption carries the cell-specific findings; the prose below signposts which group a given table belongs to and what it adjudicates.

\begin{table}[htbp]
\centering
\tiny
\setlength{\tabcolsep}{4pt}
\begin{tabular}{l l c c c c c c c c c c c}
\toprule
& & \multicolumn{2}{c}{Germanic} & \multicolumn{3}{c}{Romance} & Indic & Semitic & Sinitic & Japonic & Turkic & Aus. \\
\cmidrule(lr){3-4} \cmidrule(lr){5-7}
Scheme & Gen. & eng & nld & fra & spa & por & hin & arb & zho & jpn & tur & vie \\
\midrule
\multicolumn{13}{l}{\textit{Base regime (FLORES+ continuation prompts)}} \\
\midrule
KGW     & Mistral & $0.000$   & $+0.004$ & $+0.004$ & $+0.002$ & $0.000$ & $-0.008$ & $0.000$   & $-0.006$ & $0.000$   & $-0.004$ & $+0.002$ \\
        & Gemma   & $+0.006$ & $+0.030$ & $+0.008$ & $+0.016$ & $+0.038$ & $-0.020$ & $+0.044$ & $-0.010$ & $-0.064$ & $0.000$   & $+0.010$ \\
        & Qwen    & $+0.028$ & $+0.018$ & $-0.002$ & $-0.002$ & $+0.002$ & $-0.008$ & $0.000$   & $+0.008$ & $0.000$   & $-0.030$ & $0.000$   \\
\cmidrule(lr){1-13}
Unigram & Mistral & $+0.074$ & $+0.354$ & $+0.218$ & $+0.292$ & $+0.142$ & $-0.062$ & $+0.004$ & $+0.064$ & $-0.018$ & $+0.074$ & $-0.012$ \\
        & Gemma   & $+0.996$ & $+0.992$ & $+0.968$ & $+0.990$ & $+0.960$ & $0.000$   & $+0.958$ & $+0.976$ & $+0.984$ & $+0.960$ & $0.000$   \\
        & Qwen    & $+0.024$ & $+0.034$ & $-0.040$ & $+0.002$ & $+0.022$ & $+0.106$ & $+0.230$ & $+0.134$ & $-0.008$ & $+0.092$ & $-0.034$ \\
\cmidrule(lr){1-13}
SynthID & Mistral & $0.000$   & $0.000$   & $0.000$   & $0.000$   & $0.000$   & $0.000$   & $0.000$   & $+0.002$ & $0.000$   & $0.000$   & $0.000$   \\
        & Gemma   & $0.000$   & $0.000$   & $0.000$   & $0.000$   & $0.000$   & $0.000$   & $0.000$   & $0.000$   & $0.000$   & $0.000$   & $0.000$   \\
        & Qwen    & $0.000$   & $0.000$   & $0.000$   & $0.000$   & $0.000$   & $+0.002$ & $0.000$   & $+0.006$ & $0.000$   & $-0.004$ & $0.000$   \\
\cmidrule(lr){1-13}
EXPEdit & Mistral & $0.000$   & $+0.002$ & $0.000$   & $0.000$   & $0.000$   & $0.000$   & $0.000$   & $0.000$   & $0.000$   & $0.000$   & $0.000$   \\
        & Gemma   & $0.000$   & $0.000$   & $0.000$   & $0.000$   & $0.000$   & $0.000$   & $0.000$   & $0.000$   & $0.000$   & $0.000$   & $0.000$   \\
        & Qwen    & $+0.002$ & $-0.006$ & $0.000$   & $0.000$   & $0.000$   & $+0.012$ & $0.000$   & $0.000$   & $0.000$   & $+0.002$ & $0.000$   \\
\cmidrule(lr){1-13}
DIP     & Mistral & $0.000$   & $0.000$   & $+0.002$ & $+0.002$ & $+0.002$ & $+0.002$ & $0.000$   & $-0.014$ & $0.000$   & $+0.004$ & $0.000$   \\
        & Gemma   & $0.000$   & $-0.004$ & $+0.002$ & $0.000$   & $+0.002$ & $+0.010$ & $+0.004$ & $+0.004$ & $+0.018$ & $-0.010$ & $0.000$   \\
        & Qwen    & $0.000$   & $-0.008$ & $+0.004$ & $+0.008$ & $-0.004$ & $-0.004$ & $+0.002$ & $+0.010$ & $0.000$   & $-0.028$ & $+0.002$ \\
\cmidrule(lr){1-13}
XSIR    & Mistral & $+0.024$ & $-0.054$ & $+0.166$ & $+0.106$ & $+0.086$ & $-0.022$ & $-0.066$ & $+0.092$ & $+0.014$ & $+0.014$ & $+0.178$ \\
        & Gemma   & $+0.028$ & $+0.054$ & $+0.054$ & $+0.114$ & $+0.026$ & $+0.146$ & $+0.056$ & $-0.042$ & $+0.168$ & $+0.122$ & $-0.030$ \\
        & Qwen    & $+0.516$ & $+0.774$ & $+0.770$ & $+0.802$ & $+0.912$ & $-0.286$ & $+0.684$ & $+0.638$ & $+0.464$ & $+0.360$ & $+0.732$ \\
\midrule
\multicolumn{13}{l}{\textit{Instruct regime (Aya native-speaker prompts)}} \\
\midrule
KGW     & Mistral & $+0.112$ & $+0.032$ & $-0.250$ & $+0.028$ & $+0.118$ & $-0.016$ & $+0.014$ & $+0.030$ & $-0.018$ & $+0.012$ & $-0.004$ \\
        & Gemma   & $+0.030$ & $-0.070$ & $+0.108$ & $+0.046$ & $+0.112$ & $-0.038$ & $0.000$   & $-0.036$ & $-0.082$ & $+0.134$ & $+0.034$ \\
        & Qwen    & $+0.012$ & $+0.014$ & $-0.036$ & $-0.058$ & $+0.110$ & $-0.106$ & $+0.138$ & $+0.016$ & $+0.084$ & $+0.058$ & $+0.140$ \\
\cmidrule(lr){1-13}
Unigram & Mistral & $+0.340$ & $+0.380$ & $+0.090$ & $+0.282$ & $+0.316$ & $-0.106$ & $0.000$   & $+0.070$ & $-0.178$ & $+0.282$ & $+0.052$ \\
        & Gemma   & $+0.024$ & $-0.054$ & $+0.174$ & $+0.178$ & $+0.056$ & $-0.070$ & $-0.036$ & $+0.376$ & $+0.276$ & $+0.272$ & $+0.050$ \\
        & Qwen    & $+0.084$ & $-0.018$ & $-0.108$ & $-0.012$ & $+0.078$ & $+0.128$ & $+0.438$ & $+0.352$ & $+0.094$ & $+0.088$ & $-0.132$ \\
\cmidrule(lr){1-13}
SynthID & Mistral & $+0.056$ & $+0.020$ & $-0.026$ & $-0.034$ & $-0.048$ & $+0.016$ & $0.000$   & $0.000$   & $-0.022$ & $0.000$   & $+0.016$ \\
        & Gemma   & $+0.050$ & $-0.030$ & $+0.046$ & $-0.022$ & $-0.020$ & $-0.022$ & $-0.124$ & $+0.038$ & $-0.102$ & $+0.094$ & $+0.036$ \\
        & Qwen    & $+0.018$ & $-0.002$ & $+0.010$ & $-0.022$ & $-0.002$ & $+0.016$ & $0.000$   & $0.000$   & $0.000$   & $0.000$   & $0.000$   \\
\cmidrule(lr){1-13}
EXPEdit & Mistral & $-0.030$ & $+0.062$ & $-0.080$ & $+0.028$ & $+0.130$ & $-0.012$ & $+0.006$ & $+0.004$ & $+0.028$ & $+0.016$ & $-0.002$ \\
        & Gemma   & $+0.002$ & $+0.042$ & $-0.040$ & $0.000$   & $-0.018$ & $-0.022$ & $+0.046$ & $+0.066$ & $+0.066$ & $+0.002$ & $-0.014$ \\
        & Qwen    & $+0.046$ & $-0.002$ & $-0.072$ & $-0.006$ & $+0.050$ & $+0.048$ & $+0.032$ & $0.000$   & $+0.004$ & $-0.020$ & $+0.056$ \\
\cmidrule(lr){1-13}
DIP     & Mistral & $-0.042$ & $+0.026$ & $-0.016$ & $+0.002$ & $+0.008$ & $+0.058$ & $+0.048$ & $+0.046$ & $+0.122$ & $+0.032$ & $-0.038$ \\
        & Gemma   & $-0.016$ & $-0.014$ & $+0.010$ & $0.000$   & $+0.052$ & $+0.076$ & $-0.004$ & $+0.030$ & $+0.004$ & $-0.030$ & $0.000$   \\
        & Qwen    & $-0.046$ & $0.000$   & $-0.038$ & $0.000$   & $+0.016$ & $+0.030$ & $-0.008$ & $+0.042$ & $+0.088$ & $+0.006$ & $-0.008$ \\
\cmidrule(lr){1-13}
XSIR    & Mistral & $+0.064$ & $-0.180$ & $+0.044$ & $-0.032$ & $-0.008$ & $+0.088$ & $-0.152$ & $+0.084$ & $+0.186$ & $+0.208$ & $+0.078$ \\
        & Gemma   & $+0.110$ & $+0.062$ & $-0.008$ & $+0.046$ & $-0.022$ & $+0.026$ & $-0.024$ & $+0.174$ & $-0.006$ & $+0.116$ & $+0.002$ \\
        & Qwen    & $+0.208$ & $+0.308$ & $+0.212$ & $+0.224$ & $+0.286$ & $-0.248$ & $+0.274$ & $+0.406$ & $+0.292$ & $+0.444$ & $+0.264$ \\
\bottomrule
\end{tabular}
\caption{Per-language calibration gap $\Delta_l = \mathrm{TPR}_l(\tau_l) - \mathrm{TPR}_l(\tau_g)$, base-FLORES (top) and instruct-Aya (bottom), subset \texttt{all}, $\alpha = 0.01$. Positive: per-language calibration recovers detection. Negative: per-language calibration worsens detection. Zero: detection insensitive to calibration locality. Row means equal $-1\times$ the $\Delta$ column of Table~\ref{tab:detection_summary}.}
\label{tab:detection_calibration_gap}
\end{table}

\subsection{Per-Language Detail}\label{app:detection_perlang}
Figures~\ref{fig:detection_heatmap_base} and~\ref{fig:detection_heatmap_instruct} render the per-language TPR and AUC vectors for both regimes. The heatmap is the visual entry point to every cell-language pattern that the headline table's mean, min, and GE$_2$ statistics collapse: the Unigram-Gemma all-zero row under base, the XSIR-Qwen Hindi outlier, the EXPEdit-Gemma broad AUC degradation under instruct. Table~\ref{tab:detection_calibration_gap} then shows the per-language $\Delta_l = \mathrm{TPR}_l(\tau_l) - \mathrm{TPR}_l(\tau_g)$ matrix, decomposing the headline $\Delta$ column into its per-language sources. Together the two items establish that calibration locality is doing structurally different work in different cells: a near-noise correction of order $\pm 0.03$ on strong-detection cells, a near-total recovery on the base Unigram-Gemma row, and a smaller-scale broad recovery across most instruct cells.

\subsection{Detection-Threshold Robustness} \label{app:detection_threshold_robustness}
The calibration-failure diagnoses in \S\ref{sec:findings_detection} depend on two operating choices: the FPR target $\alpha$ and the LID-adherence subset on which the null is calibrated. Table~\ref{tab:detection_alpha_05} repeats the headline summary at $\alpha = 0.05$, and Table~\ref{tab:detection_on_target} restricts the unwatermarked null to LID-passing on-target generations. The first table addresses whether the score-cap saturation underlying Unigram-Gemma at $\alpha = 0.01$ persists at a looser FPR target; the second addresses whether that saturation is driven by off-target generations contaminating the calibration pool. Read together, the two robustness checks distinguish two distinct base-regime failure modes: Unigram-Gemma is an off-target-mixture artifact that dissolves under either operating change, while XSIR-Qwen is a deeper threshold misalignment that persists across both.

\begin{table}[htbp]
\centering
\scriptsize
\setlength{\tabcolsep}{4pt}
\begin{tabular}{l l c c c c c c c}
\toprule
& & \multicolumn{5}{c}{TPR at $\tau_g$ (global)} & \multicolumn{2}{c}{TPR at $\tau_l$ (per-lang)} \\
\cmidrule(lr){3-7} \cmidrule(lr){8-9}
Scheme & Gen. & Mean TPR & Min TPR & DI$_{<.8}$ & GE$_2$ & Btw.\,\% & Mean TPR & $\Delta$ \\
\midrule
\multicolumn{9}{l}{\textit{Base regime (FLORES+ continuation prompts)}} \\
\midrule
KGW     & Mistral & 0.997 & 0.992 & 0  & $<\!0.001$ & 61.1  & 0.997 & $+0.000$ \\
        & Gemma   & 0.991 & 0.980 & 0  & $<\!0.001$ & 93.2  & 0.993 & $-0.002$ \\
        & Qwen    & 0.987 & 0.952 & 0  & $<\!0.001$ & 57.8  & 0.986 & $+0.000$ \\
\cmidrule(lr){1-9}
Unigram & Mistral & 0.977 & 0.920 & 0  & $<\!0.001$ & 54.9  & 0.988 & $-0.011$ \\
        & Gemma   & 0.988 & 0.976 & 0  & $<\!0.001$ & 55.9  & 0.898 & $+0.089$ \\
        & Qwen    & 0.961 & 0.860 & 0  & 0.001      & 86.3  & 0.979 & $-0.018$ \\
\cmidrule(lr){1-9}
SynthID & Mistral & 1.000 & 0.998 & 0  & $<\!0.001$ & 100.0 & 1.000 & $+0.000$ \\
        & Gemma   & 1.000 & 0.998 & 0  & $<\!0.001$ & 100.0 & 1.000 & $+0.000$ \\
        & Qwen    & 0.999 & 0.996 & 0  & $<\!0.001$ & 74.3  & 0.999 & $+0.000$ \\
\cmidrule(lr){1-9}
EXPEdit & Mistral & 1.000 & 0.996 & 0  & $<\!0.001$ & 100.0 & 1.000 & $+0.000$ \\
        & Gemma   & 1.000 & 0.998 & 0  & $<\!0.001$ & 100.0 & 1.000 & $+0.000$ \\
        & Qwen    & 0.995 & 0.966 & 0  & $<\!0.001$ & 99.2  & 0.995 & $-0.000$ \\
\cmidrule(lr){1-9}
DIP     & Mistral & 0.990 & 0.972 & 0  & $<\!0.001$ & 94.7  & 0.989 & $+0.001$ \\
        & Gemma   & 0.981 & 0.942 & 0  & $<\!0.001$ & 97.9  & 0.981 & $-0.001$ \\
        & Qwen    & 0.966 & 0.896 & 0  & 0.001      & 91.6  & 0.966 & $+0.000$ \\
\cmidrule(lr){1-9}
XSIR    & Mistral & 0.924 & 0.868 & 0  & 0.001      & 83.3  & 0.935 & $-0.011$ \\
        & Gemma   & 0.929 & 0.840 & 0  & 0.001      & 92.8  & 0.944 & $-0.015$ \\
        & Qwen    & 0.503 & 0.276 & 9  & 0.076      & 94.6  & 0.919 & $-0.416$ \\
\midrule
\multicolumn{9}{l}{\textit{Instruct regime (Aya native-speaker prompts)}} \\
\midrule
KGW     & Mistral & 0.801 & 0.622 & 2  & 0.006      & 85.0  & 0.813 & $-0.012$ \\
        & Gemma   & 0.616 & 0.474 & 10 & 0.015      & 79.7  & 0.617 & $-0.001$ \\
        & Qwen    & 0.860 & 0.740 & 1  & 0.003      & 78.1  & 0.869 & $-0.009$ \\
\cmidrule(lr){1-9}
Unigram & Mistral & 0.517 & 0.090 & 5  & 0.149      & 95.8  & 0.659 & $-0.143$ \\
        & Gemma   & 0.380 & 0.154 & 5  & 0.050      & 80.3  & 0.408 & $-0.028$ \\
        & Qwen    & 0.727 & 0.366 & 4  & 0.018      & 93.8  & 0.790 & $-0.063$ \\
\cmidrule(lr){1-9}
SynthID & Mistral & 0.967 & 0.926 & 0  & $<\!0.001$ & 77.8  & 0.969 & $-0.002$ \\
        & Gemma   & 0.791 & 0.662 & 4  & 0.005      & 75.9  & 0.785 & $+0.005$ \\
        & Qwen    & 0.971 & 0.898 & 0  & $<\!0.001$ & 95.7  & 0.973 & $-0.002$ \\
\cmidrule(lr){1-9}
EXPEdit & Mistral & 0.665 & 0.354 & 4  & 0.036      & 76.7  & 0.682 & $-0.017$ \\
        & Gemma   & 0.341 & 0.126 & 10 & 0.049      & 83.3  & 0.339 & $+0.002$ \\
        & Qwen    & 0.583 & 0.230 & 9  & 0.050      & 82.5  & 0.596 & $-0.013$ \\
\cmidrule(lr){1-9}
DIP     & Mistral & 0.666 & 0.478 & 6  & 0.017      & 98.8  & 0.672 & $-0.006$ \\
        & Gemma   & 0.410 & 0.302 & 10 & 0.031      & 93.0  & 0.416 & $-0.007$ \\
        & Qwen    & 0.698 & 0.510 & 7  & 0.012      & 99.7  & 0.700 & $-0.002$ \\
\cmidrule(lr){1-9}
XSIR    & Mistral & 0.510 & 0.256 & 9  & 0.051      & 86.4  & 0.523 & $-0.013$ \\
        & Gemma   & 0.267 & 0.072 & 10 & 0.092      & 94.3  & 0.300 & $-0.033$ \\
        & Qwen    & 0.169 & 0.024 & 10 & 0.811      & 97.5  & 0.513 & $-0.344$ \\
\bottomrule
\end{tabular}
\caption{Detection fairness summary at $\alpha = 0.05$, base-FLORES (top) and instruct-Aya (bottom), subset \texttt{all}. Format mirrors Table~\ref{tab:detection_summary}; AUC columns are omitted (threshold-independent, identical to Table~\ref{tab:detection_summary}) and bootstrap CIs on Mean TPR are omitted for compactness. $L = 11$ throughout. Values below $10^{-3}$ reported as $<\!0.001$.}
\label{tab:detection_alpha_05}
\end{table}

\begin{table}[htbp]
\centering
\scriptsize
\setlength{\tabcolsep}{4pt}
\begin{tabular}{l l c c c c c c c c}
\toprule
& & \multicolumn{4}{c}{Base regime (FLORES+ continuation)} & \multicolumn{4}{c}{Instruct regime (Aya native-speaker)} \\
\cmidrule(lr){3-6} \cmidrule(lr){7-10}
Scheme & Gen. & Mean TPR & Min TPR & DI$_{<.8}$ & $\Delta$ vs.\ \texttt{all} & Mean TPR & Min TPR & DI$_{<.8}$ & $\Delta$ vs.\ \texttt{all} \\
\midrule
KGW     & Mistral & 0.995 & 0.983 & 0  & $+0.003$ & 0.681 & 0.425 & 6  & $+0.055$ \\
        & Gemma   & 0.973 & 0.939 & 0  & $+0.004$ & 0.426 & 0.264 & 10 & $+0.037$ \\
        & Qwen    & 0.987 & 0.963 & 0  & $+0.022$ & 0.697 & 0.520 & 8  & $+0.008$ \\
\cmidrule(lr){1-10}
Unigram & Mistral & 0.885 & 0.654 & 2  & $+0.017$ & 0.388 & 0.011 & 7  & $+0.077$ \\
        & Gemma   & 0.987 & 0.974 & 0  & $+0.987$ & 0.020 & 0.000 & 9  & $-0.017$ \\
        & Qwen    & 0.925 & 0.762 & 1  & $+0.027$ & 0.547 & 0.176 & 4  & $+0.001$ \\
\cmidrule(lr){1-10}
SynthID & Mistral & 1.000 & 0.998 & 0  & $0.000$  & 0.924 & 0.774 & 1  & $+0.019$ \\
        & Gemma   & 1.000 & 0.998 & 0  & $0.000$  & 0.610 & 0.426 & 10 & $+0.008$ \\
        & Qwen    & 0.999 & 0.996 & 0  & $+0.002$ & 0.941 & 0.764 & 1  & $0.000$  \\
\cmidrule(lr){1-10}
EXPEdit & Mistral & 1.000 & 1.000 & 0  & $+0.001$ & 0.675 & 0.234 & 5  & $+0.098$ \\
        & Gemma   & 1.000 & 0.997 & 0  & $0.000$  & 0.232 & 0.091 & 10 & $+0.007$ \\
        & Qwen    & 0.994 & 0.941 & 0  & $+0.003$ & 0.508 & 0.059 & 10 & $+0.012$ \\
\cmidrule(lr){1-10}
DIP     & Mistral & 0.980 & 0.952 & 0  & $+0.005$ & 0.488 & 0.266 & 9  & $+0.033$ \\
        & Gemma   & 0.977 & 0.954 & 0  & $+0.014$ & 0.198 & 0.105 & 10 & $+0.002$ \\
        & Qwen    & 0.957 & 0.848 & 0  & $+0.026$ & 0.507 & 0.200 & 8  & $-0.006$ \\
\cmidrule(lr){1-10}
XSIR    & Mistral & 0.799 & 0.688 & 2  & $+0.015$ & 0.296 & 0.081 & 10 & $+0.030$ \\
        & Gemma   & 0.793 & 0.600 & 3  & $+0.027$ & 0.092 & 0.019 & 10 & $+0.011$ \\
        & Qwen    & 0.207 & 0.047 & 10 & $-0.008$ & 0.047 & 0.000 & 10 & $+0.004$ \\
\bottomrule
\end{tabular}
\caption{Detection summary on the on-target subset, $\alpha = 0.01$, global threshold $\tau_g$, base-FLORES (left) and instruct-Aya (right). Thresholds recalibrated on each cell's on-target unwatermarked pool; $L = 11$ throughout. The off-target subset is omitted: in the base regime no cell meets the $n_{\text{neg}} \geq 200$ inclusion floor, and in the instruct regime only six (Mistral-generator) cells pass the relaxed $\alpha = 0.05$ floor with $L \leq 3$, insufficient for the fairness aggregates this appendix reports. The $\Delta$ vs.\ \texttt{all} columns report change in Mean TPR relative to subset \texttt{all} (Table~\ref{tab:detection_summary}).}
\label{tab:detection_on_target}
\end{table}

\subsection{Floor-Statistic Robustness}\label{app:detection_disparity_robustness}
The headline table reports Min TPR as a worst-language floor, which is sensitive to isolated single-language outliers. Table~\ref{tab:bottom_quintile} replaces Min TPR with the mean of the bottom two languages (the bottom quintile at $L=11$), testing whether the headline minima are driven by isolated outliers or by low-performing clusters. The bottom-quintile mean and Min TPR identify the same worst-performing cells in both regimes, with the bottom-quintile values typically within $0.05$ of Min TPR on cells where the floor is genuine and substantially higher only on the handful of cells where a single language sits well below the rest. The check supports the headline floor statistic rather than displacing it.

\begin{table}[htbp]
\centering
\scriptsize
\setlength{\tabcolsep}{4pt}
\begin{tabular}{l l c c l c c l}
\toprule
& & \multicolumn{3}{c}{Base regime (FLORES+)} & \multicolumn{3}{c}{Instruct regime (Aya)} \\
\cmidrule(lr){3-5} \cmidrule(lr){6-8}
Scheme & Gen. & Min TPR & Bot.-q mean & Bot.-q langs & Min TPR & Bot.-q mean & Bot.-q langs \\
\midrule
KGW     & Mistral & 0.978 & 0.979 & \texttt{zho, hin} & 0.386 & 0.447 & \texttt{fra, por} \\
        & Gemma   & 0.934 & 0.943 & \texttt{arb, por} & 0.222 & 0.251 & \texttt{por, vie} \\
        & Qwen    & 0.880 & 0.894 & \texttt{tur, nld} & 0.510 & 0.564 & \texttt{por, nld} \\
\cmidrule(lr){1-8}
Unigram & Mistral & 0.638 & 0.672 & \texttt{nld, spa} & 0.022 & 0.029 & \texttt{spa, fra} \\
        & Gemma   & 0.000 & 0.000 & \texttt{eng, fra} & 0.010 & 0.010 & \texttt{por, tur} \\
        & Qwen    & 0.722 & 0.744 & \texttt{tur, arb} & 0.210 & 0.306 & \texttt{arb, hin} \\
\cmidrule(lr){1-8}
SynthID & Mistral & 0.998 & 0.998 & \texttt{zho, tur} & 0.782 & 0.788 & \texttt{fra, spa} \\
        & Gemma   & 0.998 & 0.999 & \texttt{vie, eng} & 0.422 & 0.437 & \texttt{vie, por} \\
        & Qwen    & 0.990 & 0.992 & \texttt{tur, zho} & 0.792 & 0.842 & \texttt{hin, por} \\
\cmidrule(lr){1-8}
EXPEdit & Mistral & 0.996 & 0.997 & \texttt{zho, nld} & 0.238 & 0.244 & \texttt{eng, spa} \\
        & Gemma   & 0.998 & 0.998 & \texttt{jpn, vie} & 0.084 & 0.101 & \texttt{jpn, por} \\
        & Qwen    & 0.946 & 0.968 & \texttt{hin, nld} & 0.174 & 0.226 & \texttt{hin, eng} \\
\cmidrule(lr){1-8}
DIP     & Mistral & 0.938 & 0.950 & \texttt{zho, jpn} & 0.264 & 0.265 & \texttt{fra, spa} \\
        & Gemma   & 0.914 & 0.917 & \texttt{hin, vie} & 0.110 & 0.115 & \texttt{por, jpn} \\
        & Qwen    & 0.798 & 0.819 & \texttt{tur, hin} & 0.296 & 0.345 & \texttt{hin, fra} \\
\cmidrule(lr){1-8}
XSIR    & Mistral & 0.672 & 0.673 & \texttt{zho, vie} & 0.088 & 0.116 & \texttt{fra, spa} \\
        & Gemma   & 0.588 & 0.635 & \texttt{jpn, hin} & 0.016 & 0.019 & \texttt{tur, jpn} \\
        & Qwen    & 0.050 & 0.064 & \texttt{por, fra} & 0.000 & 0.000 & \texttt{nld, vie} \\
\bottomrule
\end{tabular}
\caption{Bottom-quintile-mean Rawlsian floor as a stability companion to Min TPR (Table~\ref{tab:detection_summary}), base-FLORES (left) and instruct-Aya (right), subset \texttt{all}, $\alpha = 0.01$, global threshold $\tau_g$. With $L = 11$ evaluation languages the bottom quintile is $\lfloor 0.2 \cdot 11 \rfloor = 2$ languages; the columns report the mean TPR of the two worst-performing languages per cell, with the languages listed minimum first (arbitrary tie-breaking). The Unigram-Gemma base cell is degenerate (all eleven per-language TPRs are zero); its bottom-quintile language identities are reported but uninformative.}
\label{tab:bottom_quintile}
\end{table}

\subsection{Decomposition Robustness} \label{app:detection_decomposition_robustness}
The family-structure claim of \S\ref{sec:findings_detection} rests on a single decomposition: GE$_2$ over the typological-family partition. Table~\ref{tab:ge_0_robustness} repeats it under GE$_0$ (mean log deviation, more sensitive to inequality at the low end of the per-language vector), and Table~\ref{tab:ge_partition_robustness} repeats it under a script-based partition. Between-family share remains above 55\% in every well-defined cell under both inequality measures across both regimes, though \S\ref{app:sensitivity} shows the chance level is itself near 70\%. The family partition consistently captures more between-group variance than the script partition for cells with non-trivial total disparity, locating the structural component at the typology level rather than the writing-system level.

\begin{table}[htbp]
\centering
\scriptsize
\setlength{\tabcolsep}{4pt}
\begin{tabular}{l l c c c c c c c c}
\toprule
& & \multicolumn{4}{c}{Base regime (FLORES+)} & \multicolumn{4}{c}{Instruct regime (Aya)} \\
\cmidrule(lr){3-6} \cmidrule(lr){7-10}
& & \multicolumn{2}{c}{$\mathrm{GE}_2$} & \multicolumn{2}{c}{$\mathrm{GE}_0$} & \multicolumn{2}{c}{$\mathrm{GE}_2$} & \multicolumn{2}{c}{$\mathrm{GE}_0$} \\
\cmidrule(lr){3-4} \cmidrule(lr){5-6} \cmidrule(lr){7-8} \cmidrule(lr){9-10}
Scheme & Gen. & Total & Btw.\,\% & Total & Btw.\,\% & Total & Btw.\,\% & Total & Btw.\,\% \\
\midrule
KGW     & Mistral & $<\!0.001$ & 83.6  & $<\!0.001$       & 83.7         & 0.026 & 95.3 & 0.027 & 91.4 \\
        & Gemma   & $<\!0.001$ & 72.4  & $<\!0.001$       & 72.7         & 0.053 & 82.4 & 0.046 & 73.3 \\
        & Qwen    & $<\!0.001$      & 85.1  & 0.001            & 85.1         & 0.011 & 78.6 & 0.011 & 72.5 \\
\cmidrule(lr){1-10}
Unigram & Mistral & 0.008      & 61.3  & 0.009            & 55.7         & 0.297 & 99.8 & 0.556 & 94.8 \\
        & Gemma   & ---        & ---   & $<\!0.001^\dagger$ & $100.0^\dagger$ & 0.226 & 81.2 & 0.243 & 73.9 \\
        & Qwen    & 0.004      & 93.8  & 0.005            & 94.6         & 0.032 & 91.7 & 0.046 & 94.6 \\
\cmidrule(lr){1-10}
SynthID & Mistral & $<\!0.001$ & 100.0 & $<\!0.001$       & 100.0        & 0.003 & 77.7 & 0.003 & 75.6 \\
        & Gemma   & $<\!0.001$ & 100.0 & $<\!0.001$       & 100.0        & 0.020 & 79.7 & 0.019 & 75.6 \\
        & Qwen    & $<\!0.001$ & 97.5  & $<\!0.001$       & 97.5         & 0.002 & 91.9 & 0.002 & 92.2 \\
\cmidrule(lr){1-10}
EXPEdit & Mistral & $<\!0.001$ & 88.0  & $<\!0.001$       & 88.1         & 0.067 & 80.0 & 0.087 & 68.2 \\
        & Gemma   & $<\!0.001$ & 100.0 & $<\!0.001$       & 100.0        & 0.071 & 84.2 & 0.077 & 79.0 \\
        & Qwen    & $<\!0.001$ & 99.2  & $<\!0.001$       & 99.2         & 0.080 & 84.4 & 0.093 & 81.4 \\
\cmidrule(lr){1-10}
DIP     & Mistral & $<\!0.001$ & 92.6  & $<\!0.001$       & 92.8         & 0.058 & 99.5 & 0.059 & 99.2 \\
        & Gemma   & $<\!0.001$ & 99.5  & $<\!0.001$       & 99.5         & 0.084 & 91.6 & 0.068 & 84.4 \\
        & Qwen    & 0.003      & 94.7  & 0.003            & 94.6         & 0.035 & 99.6 & 0.035 & 99.6 \\
\cmidrule(lr){1-10}
XSIR    & Mistral & 0.004      & 91.5  & 0.004            & 92.1         & 0.185 & 91.8 & 0.160 & 83.4 \\
        & Gemma   & 0.006      & 96.0  & 0.007            & 96.3         & 0.175 & 76.5 & 0.220 & 81.1 \\
        & Qwen    & 0.493      & 98.9  & 0.326            & 92.5         & 3.787 & 99.9 & 3.244 & 79.0 \\
\bottomrule
\end{tabular}
\caption{Robustness of the generalized-entropy decomposition to choice of inequality measure: $\mathrm{GE}_2$ (half the squared coefficient of variation, more sensitive to inequality at the top of the per-language TPR distribution) versus $\mathrm{GE}_0$ (mean log deviation, more sensitive to inequality at the bottom). Base-FLORES (left) and instruct-Aya (right), subset \texttt{all}, $\alpha = 0.01$, global threshold $\tau_g$; typological partition as in Table~\ref{tab:detection_summary}. Total values below $10^{-3}$ reported as $<\!0.001$. The Unigram-Gemma base $\mathrm{GE}_0$ entry is obtained by replacing zero TPRs with a small $\epsilon$ (per the strict-positivity requirement of $\mathrm{GE}_0$); the resulting $\mathrm{Total} = 0$ and $\mathrm{Btw.\,\%} = 100.0$ are regularization artifacts, marked $^\dagger$.}
\label{tab:ge_0_robustness}
\end{table}

\begin{table}[htbp]
\centering
\scriptsize
\setlength{\tabcolsep}{4pt}
\begin{tabular}{l l c c c c c c}
\toprule
& & \multicolumn{3}{c}{Base regime (FLORES+)} & \multicolumn{3}{c}{Instruct regime (Aya)} \\
\cmidrule(lr){3-5} \cmidrule(lr){6-8}
& & & \multicolumn{2}{c}{Btw.\,\%} & & \multicolumn{2}{c}{Btw.\,\%} \\
\cmidrule(lr){4-5} \cmidrule(lr){7-8}
Scheme & Gen. & $\mathrm{GE}_2$ & Family & Script & $\mathrm{GE}_2$ & Family & Script \\
\midrule
KGW     & Mistral & $<\!0.001$ & 83.6  & 46.2 & 0.026 & 95.3 & 32.8 \\
        & Gemma   & $<\!0.001$ & 72.4  & 62.2 & 0.053 & 82.4 & 29.8 \\
        & Qwen    & $<\!0.001$ & 85.1  & 26.2 & 0.011 & 78.6 & 26.7 \\
\cmidrule(lr){1-8}
Unigram & Mistral & 0.008      & 61.3  & 28.8 & 0.297 & 99.8 & 46.3 \\
        & Gemma   & ---        & ---   & ---  & 0.226 & 81.2 & 61.7 \\
        & Qwen    & 0.004      & 93.8  & 26.1 & 0.032 & 91.7 & 75.0 \\
\cmidrule(lr){1-8}
SynthID & Mistral & $<\!0.001$ & 100.0 & 17.1 & 0.003 & 77.7 & 33.0 \\
        & Gemma   & $<\!0.001$ & 100.0 & 5.7  & 0.020 & 79.7 & 47.1 \\
        & Qwen    & $<\!0.001$ & 97.5  & 7.5  & 0.002 & 91.9 & 79.3 \\
\cmidrule(lr){1-8}
EXPEdit & Mistral & $<\!0.001$ & 88.0  & 31.7 & 0.067 & 80.0 & 30.2 \\
        & Gemma   & $<\!0.001$ & 100.0 & 17.1 & 0.071 & 84.2 & 8.5  \\
        & Qwen    & $<\!0.001$ & 99.2  & 95.8 & 0.080 & 84.4 & 41.6 \\
\cmidrule(lr){1-8}
DIP     & Mistral & $<\!0.001$ & 92.6  & 69.3 & 0.058 & 99.5 & 38.9 \\
        & Gemma   & $<\!0.001$ & 99.5  & 48.2 & 0.084 & 91.6 & 25.2 \\
        & Qwen    & 0.003      & 94.7  & 32.5 & 0.035 & 99.6 & 53.4 \\
\cmidrule(lr){1-8}
XSIR    & Mistral & 0.004      & 91.5  & 54.8 & 0.185 & 91.8 & 64.7 \\
        & Gemma   & 0.006      & 96.0  & 12.7 & 0.175 & 76.5 & 7.6  \\
        & Qwen    & 0.493      & 98.9  & 93.5 & 3.787 & 99.9 & 99.8 \\
\bottomrule
\end{tabular}
\caption{Robustness of the $\mathrm{GE}_2$ between-share to choice of partition: typological family (8 groups: Germanic, Romance, Indic, Semitic, Sinitic, Japonic, Turkic, Austroasiatic, with 5 of 11 languages in non-singleton groups) vs.\ script (4 groups: Latin pooling eng, nld, fra, spa, por, tur, vie; Devanagari$=$hin; Arabic$=$arb; Han pooling zho, jpn; with 9 of 11 languages in non-singleton groups). Base-FLORES (left) and instruct-Aya (right), subset \texttt{all}, $\alpha = 0.01$, global threshold $\tau_g$. $\mathrm{GE}_2$ Total depends only on the per-language vector and is partition-independent (shown once per regime). Totals below $10^{-3}$ reported as $<\!0.001$; for these cells the between-share values involve ratios of near-zero variance components and should not be over-interpreted.}
\label{tab:ge_partition_robustness}
\end{table}

\subsection{Partition-Shape Sensitivity}\label{app:sensitivity}
Six of the eight families are singletons, so a between-family share is high even under a random labeling. The shares in Tables~\ref{tab:detection_summary} and \ref{tab:q_5_2_summary} are read against an exact null of the same shape. Holding every per-language score fixed, we apply each of the 4{,}620 relabelings of the eleven languages that preserve the family shape (one three-member group, one two-member group, six singletons) jointly to every cell with GE$_2$ above the $10^{-3}$ floor. The share of relabelings whose grid-mean between-family share matches or exceeds the observed one is an exact $p$-value, which we Holm-correct over the four strata. The null centers near 70\% by construction, since for this shape the expected within-family share of a random labeling is $3/10$. Table~\ref{tab:shape_null} shows detection clears the null in both regimes, and in the base regime only 10 of the 4{,}620 relabelings match the observed mean. MAUVE sits at chance in the base regime and clears the null in the instruct regime only before correction. On the on-target subset the detection result holds with exact $p = .0035$ and $.0067$, which the Holm correction can raise to at most $.014$ and $.027$.

\begin{table}[htbp]
\centering
\scriptsize
\begin{tabular}{lcccc}
\toprule
Stratum & Cells & Observed mean share (\%) & Null median (95th pct.) & Exact $p$ (Holm) \\
\midrule
Detection, base     & 6  & 89.4 & 69.8 (82.7) & .0022 (.0087) \\
Detection, instruct & 18 & 88.1 & 70.0 (83.7) & .011 (.033) \\
MAUVE, base         & 14 & 69.3 & 71.6 (88.1) & .57 (.57) \\
MAUVE, instruct     & 5  & 90.8 & 68.9 (90.0) & .039 (.078) \\
\bottomrule
\end{tabular}
\caption{Exact same-shape relabeling test of the between-family share, subset \texttt{all}, $\alpha = 0.01$, global threshold $\tau_g$. Cells counts the cells with GE$_2 \geq 10^{-3}$. The null is the distribution of the grid-mean between-family share over all relabelings that preserve the family shape, and $p$ is the share of relabelings at or above the observed mean, with the Holm correction over the four strata in parentheses.}
\label{tab:shape_null}
\end{table}

Restricting the decomposition to the two multi-member families (Germanic: eng, nld; Romance: fra, spa, por) gives between-family medians of 36.3, 24.4, 22.1, and 14.6 in the stratum order of Table~\ref{tab:shape_null}, against a chance level of about 25\% for that two-family shape, so the decomposition does not manufacture family structure where the typological contrast is small. When the family partition is replaced with partitions that have no singletons, on the instruct detection stratum, a coarse Indo-European-versus-rest split (eng, nld, fra, spa, por, hin against the remaining five; 462 relabelings) clears its own exact null at $p < .05$ in 8 of 18 cells, whereas a Latin-versus-non-Latin split (330 relabelings) does so in 1 of 18, consistent with the family-over-script ordering of Table~\ref{tab:ge_partition_robustness}. Dropping one language at a time changes no stratum-level conclusion; the one correction it produced is that the XSIR-Qwen instruct cell (GE$_2 = 3.787$) is driven by Hindi alone, as stated in \S\ref{sec:findings_detection}.

\clearpage

\section{Quality: Per-Paradigm Detail and Robustness}\label{qulity-app}

\subsection{Per-Cell Quality Fairness Summary}
\label{app:quality-summary}

\begin{table}[htbp]
\centering
\tiny
\setlength{\tabcolsep}{3pt}
\resizebox{\textwidth}{!}{%
\begin{tabular}{ll|rrrr|r|rrrr}
\toprule
 &  & \multicolumn{4}{c}{MAUVE} & \multicolumn{1}{c}{BERTScore F1 (rescaled)} & \multicolumn{4}{c}{PPL preservation (XGLM-7.5B)} \\
\cmidrule(lr){3-6} \cmidrule(lr){7-7} \cmidrule(lr){8-11}
Scheme & Gen & mean (min) & DI$<$.8 & GE$_2$ & Btw.\,\% & mean (min) & mean (min) & DI$<$.8 & GE$_2$ & Btw.\,\% \\
\midrule
\multicolumn{11}{l}{\textit{Base regime (FLORES+ continuation)}} \\
\midrule
KGW & Mistral & 0.947 (0.879)\textsuperscript{jpn} & 0 & $<\!0.001$ & 86.0 & -0.041 (-0.079)\textsuperscript{vie} & 0.866 (0.793)\textsuperscript{spa} & 0 & 0.002 & 66.4 \\
KGW & Gemma & 0.855 (0.435)\textsuperscript{arb} & 1 & 0.013 & 98.9 & -0.081 (-0.194)\textsuperscript{vie} & 0.899 (0.812)\textsuperscript{fra} & 0 & 0.002 & 91.0 \\
KGW & Qwen & 0.889 (0.471)\textsuperscript{hin} & 1 & 0.012 & 95.6 & -0.034 (-0.091)\textsuperscript{tur} & 0.834 (0.686)\textsuperscript{hin} & 3 & 0.005 & 94.0 \\
Unigram & Mistral & 0.747 (0.420)\textsuperscript{arb} & 5 & 0.026 & 49.5 & -0.070 (-0.134)\textsuperscript{arb} & 0.872 (0.811)\textsuperscript{spa} & 0 & 0.002 & 96.4 \\
Unigram & Gemma & 0.638 (0.444)\textsuperscript{spa} & 6 & 0.017 & 67.8 & -0.098 (-0.191)\textsuperscript{jpn} & 0.911 (0.815)\textsuperscript{nld} & 0 & 0.002 & 68.3 \\
Unigram & Qwen & 0.734 (0.123)\textsuperscript{hin} & 3 & 0.043 & 97.2 & -0.055 (-0.128)\textsuperscript{zho} & 0.896 (0.647)\textsuperscript{hin} & 1 & 0.005 & 98.0 \\
SynthID & Mistral & 0.187 (0.020)\textsuperscript{arb} & 10 & 0.435 & 59.0 & -0.115 (-0.163)\textsuperscript{vie} & 0.299 (0.145)\textsuperscript{arb} & 7 & 0.048 & 88.4 \\
SynthID & Gemma & 0.095 (0.029)\textsuperscript{tur} & 10 & 0.641 & 43.0 & -0.146 (-0.254)\textsuperscript{vie} & 0.269 (0.195)\textsuperscript{tur} & 7 & 0.021 & 73.2 \\
SynthID & Qwen & 0.183 (0.035)\textsuperscript{vie} & 10 & 0.455 & 53.0 & -0.112 (-0.188)\textsuperscript{tur} & 0.308 (0.203)\textsuperscript{arb} & 9 & 0.028 & 78.2 \\
EXPEdit & Mistral & 0.079 (0.006)\textsuperscript{arb} & 10 & 0.536 & 57.1 & -0.149 (-0.228)\textsuperscript{vie} & 0.218 (0.079)\textsuperscript{arb} & 6 & 0.085 & 99.1 \\
EXPEdit & Gemma & 0.013 (0.006)\textsuperscript{jpn} & 10 & 0.574 & 43.4 & -0.220 (-0.333)\textsuperscript{vie} & 0.135 (0.076)\textsuperscript{zho} & 9 & 0.033 & 82.7 \\
EXPEdit & Qwen & 0.067 (0.013)\textsuperscript{arb} & 10 & 0.874 & 47.9 & -0.145 (-0.245)\textsuperscript{tur} & 0.224 (0.135)\textsuperscript{tur} & 9 & 0.042 & 74.6 \\
DIP & Mistral & 0.973 (0.933)\textsuperscript{hin} & 0 & 0.000 & 95.6 & -0.015 (-0.040)\textsuperscript{vie} & 0.809 (0.750)\textsuperscript{hin} & 0 & 0.001 & 77.6 \\
DIP & Gemma & 0.932 (0.878)\textsuperscript{jpn} & 0 & $<\!0.001$ & 94.7 & -0.053 (-0.183)\textsuperscript{vie} & 0.774 (0.702)\textsuperscript{nld} & 0 & 0.001 & 68.0 \\
DIP & Qwen & 0.947 (0.864)\textsuperscript{hin} & 0 & $<\!0.001$ & 81.6 & -0.008 (-0.071)\textsuperscript{tur} & 0.811 (0.714)\textsuperscript{tur} & 0 & 0.002 & 92.7 \\
XSIR & Mistral & 0.859 (0.642)\textsuperscript{hin} & 2 & 0.006 & 86.6 & -0.062 (-0.110)\textsuperscript{hin} & 0.922 (0.816)\textsuperscript{nld} & 0 & 0.002 & 73.2 \\
XSIR & Gemma & 0.683 (0.461)\textsuperscript{fra} & 6 & 0.018 & 80.9 & -0.123 (-0.226)\textsuperscript{vie} & 0.925 (0.785)\textsuperscript{nld} & 1 & 0.002 & 75.8 \\
XSIR & Qwen & 0.850 (0.603)\textsuperscript{hin} & 2 & 0.008 & 89.9 & -0.042 (-0.129)\textsuperscript{tur} & 0.883 (0.745)\textsuperscript{hin} & 1 & 0.003 & 86.4 \\
\midrule
\multicolumn{11}{l}{\textit{Instruct regime (Aya native-speaker)}} \\
\midrule
KGW & Mistral & 0.989 (0.970)\textsuperscript{arb} & 0 & 0.000 & 58.9 & 0.255 (0.088)\textsuperscript{arb} & 0.892 (0.853)\textsuperscript{jpn} & 0 & 0.001 & 91.0 \\
KGW & Gemma & 0.990 (0.959)\textsuperscript{arb} & 0 & 0.000 & 65.3 & 0.397 (0.338)\textsuperscript{tur} & 0.967 (0.941)\textsuperscript{hin} & 0 & 0.000 & 80.0 \\
KGW & Qwen & 0.987 (0.954)\textsuperscript{hin} & 0 & 0.000 & 90.8 & 0.300 (0.192)\textsuperscript{tur} & 0.885 (0.749)\textsuperscript{hin} & 1 & 0.002 & 92.5 \\
Unigram & Mistral & 0.981 (0.938)\textsuperscript{nld} & 0 & 0.000 & 40.3 & 0.240 (0.087)\textsuperscript{arb} & 0.897 (0.835)\textsuperscript{hin} & 0 & 0.001 & 97.4 \\
Unigram & Gemma & 0.996 (0.991)\textsuperscript{jpn} & 0 & 0.000 & 54.1 & 0.400 (0.329)\textsuperscript{tur} & 0.965 (0.941)\textsuperscript{nld} & 0 & 0.000 & 78.0 \\
Unigram & Qwen & 0.980 (0.927)\textsuperscript{hin} & 0 & 0.000 & 97.7 & 0.297 (0.184)\textsuperscript{tur} & 0.885 (0.700)\textsuperscript{hin} & 1 & 0.003 & 95.9 \\
SynthID & Mistral & 0.721 (0.066)\textsuperscript{arb} & 4 & 0.107 & 99.6 & 0.149 (-0.117)\textsuperscript{arb} & 0.595 (0.185)\textsuperscript{arb} & 7 & 0.070 & 95.4 \\
SynthID & Gemma & 0.994 (0.979)\textsuperscript{fra} & 0 & 0.000 & 50.0 & 0.365 (0.296)\textsuperscript{tur} & 0.931 (0.885)\textsuperscript{zho} & 0 & 0.000 & 90.0 \\
SynthID & Qwen & 0.833 (0.237)\textsuperscript{tur} & 2 & 0.033 & 97.1 & 0.219 (0.047)\textsuperscript{tur} & 0.645 (0.442)\textsuperscript{tur} & 5 & 0.014 & 90.7 \\
EXPEdit & Mistral & 0.480 (0.023)\textsuperscript{arb} & 10 & 0.198 & 86.6 & 0.082 (-0.169)\textsuperscript{arb} & 0.549 (0.198)\textsuperscript{arb} & 7 & 0.070 & 95.7 \\
EXPEdit & Gemma & 0.840 (0.598)\textsuperscript{arb} & 3 & 0.012 & 90.0 & 0.319 (0.265)\textsuperscript{jpn} & 0.742 (0.669)\textsuperscript{arb} & 0 & 0.001 & 78.7 \\
EXPEdit & Qwen & 0.699 (0.170)\textsuperscript{tur} & 4 & 0.063 & 80.6 & 0.174 (0.016)\textsuperscript{tur} & 0.651 (0.437)\textsuperscript{tur} & 5 & 0.014 & 99.1 \\
DIP & Mistral & 0.992 (0.956)\textsuperscript{arb} & 0 & 0.000 & 96.6 & 0.259 (0.108)\textsuperscript{arb} & 0.937 (0.836)\textsuperscript{arb} & 0 & 0.001 & 94.6 \\
DIP & Gemma & 0.997 (0.991)\textsuperscript{fra} & 0 & 0.000 & 52.2 & 0.403 (0.328)\textsuperscript{tur} & 0.982 (0.946)\textsuperscript{nld} & 0 & 0.000 & 61.7 \\
DIP & Qwen & 0.994 (0.985)\textsuperscript{fra} & 0 & 0.000 & 28.0 & 0.318 (0.210)\textsuperscript{tur} & 0.925 (0.853)\textsuperscript{tur} & 0 & 0.001 & 94.6 \\
XSIR & Mistral & 0.991 (0.975)\textsuperscript{arb} & 0 & 0.000 & 91.9 & 0.258 (0.112)\textsuperscript{arb} & 0.928 (0.876)\textsuperscript{hin} & 0 & 0.000 & 97.7 \\
XSIR & Gemma & 0.995 (0.975)\textsuperscript{fra} & 0 & 0.000 & 31.1 & 0.396 (0.333)\textsuperscript{tur} & 0.979 (0.969)\textsuperscript{tur} & 0 & 0.000 & 44.6 \\
XSIR & Qwen & 0.991 (0.980)\textsuperscript{tur} & 0 & 0.000 & 90.6 & 0.311 (0.197)\textsuperscript{tur} & 0.902 (0.826)\textsuperscript{tur} & 0 & 0.001 & 84.3 \\
\bottomrule
\end{tabular}%
}
\caption{Quality fairness summary by scheme, generator, regime, and measurement, subset \texttt{all}. Each row reports mean per-language preservation with the per-cell minimum and floor language (ISO 639-3 superscript) in parentheses. DI$<$.8 counts languages with per-language preservation below $0.8 \times \max_l$. GE$_2$ uses the typological partition (Germanic, Romance, Indic, Semitic, Sinitic, Japonic, Turkic, Austroasiatic); Btw.\,\% is its between-family share (chance level ${\approx}70\%$, \S\ref{subsec:ge}). MAUVE GE$_2$ below $10^{-3}$ is shown as $<\!0.001$ where it would otherwise round to 0.001. BERTScore mean (min) is reported on rescaled F1~\cite{zhang2019bertscore}; DI$<$.8 and GE$_2$ are omitted for BERTScore because neither registers signal on the bounded raw scale and both behave pathologically on the rescaled scale.}\label{tab:q_5_2_summary}
\end{table}

\begin{table}[t]
\centering
\scriptsize
\setlength{\tabcolsep}{4pt}
\begin{tabular}{ll ccc}
\toprule
Scheme & Gen. & MAUVE & BERTScore F1 (rescaled) & PPL preservation (XGLM-7.5B) \\
\midrule
\multicolumn{5}{l}{\textit{Base regime (FLORES+ continuation)}} \\
\midrule
KGW     & Mistral & 0.947 [0.928, 0.964]     & -0.041 [-0.055, -0.027]  & 0.866 [0.832, 0.899]     \\
        & Gemma   & 0.855 [0.760, 0.920]     & -0.081 [-0.111, -0.053]  & 0.899 [0.864, 0.932]     \\
        & Qwen    & 0.889 [0.794, 0.948]     & -0.034 [-0.049, -0.019]  & 0.834 [0.785, 0.883]     \\
\cmidrule(lr){1-5}
Unigram & Mistral & 0.747 [0.647, 0.839]     & -0.070 [-0.090, -0.048]  & 0.872 [0.842, 0.905]     \\
        & Gemma   & 0.638 [0.568, 0.701]     & -0.098 [-0.122, -0.078]  & 0.911 [0.879, 0.945]     \\
        & Qwen    & 0.734 [0.590, 0.837]     & -0.055 [-0.075, -0.037]  & 0.896 [0.839, 0.936]     \\
\cmidrule(lr){1-5}
SynthID & Mistral & 0.187 [0.094, 0.292]     & -0.115 [-0.136, -0.093]  & 0.299 [0.245, 0.354]     \\
        & Gemma   & 0.095 [0.050, 0.166]     & -0.146 [-0.173, -0.124]  & 0.269 [0.238, 0.302]     \\
        & Qwen    & 0.183 [0.097, 0.289]     & -0.112 [-0.134, -0.092]  & 0.308 [0.268, 0.351]     \\
\cmidrule(lr){1-5}
EXPEdit & Mistral & 0.079 [0.036, 0.130]     & -0.149 [-0.178, -0.120]  & 0.218 [0.162, 0.273]     \\
        & Gemma   & 0.013 [0.008, 0.022]     & -0.220 [-0.252, -0.192]  & 0.135 [0.114, 0.156]     \\
        & Qwen    & 0.067 [0.029, 0.122]     & -0.145 [-0.171, -0.119]  & 0.224 [0.190, 0.263]     \\
\cmidrule(lr){1-5}
DIP     & Mistral & 0.973 [0.961, 0.983]     & -0.015 [-0.024, -0.005]  & 0.809 [0.789, 0.831]     \\
        & Gemma   & 0.932 [0.911, 0.952]     & -0.053 [-0.086, -0.029]  & 0.774 [0.750, 0.797]     \\
        & Qwen    & 0.947 [0.924, 0.967]     & -0.008 [-0.023, 0.005]   & 0.811 [0.777, 0.841]     \\
\cmidrule(lr){1-5}
XSIR    & Mistral & 0.859 [0.801, 0.912]     & -0.062 [-0.080, -0.042]  & 0.922 [0.884, 0.957]     \\
        & Gemma   & 0.683 [0.606, 0.754]     & -0.123 [-0.150, -0.104]  & 0.925 [0.886, 0.959]     \\
        & Qwen    & 0.850 [0.780, 0.908]     & -0.042 [-0.065, -0.020]  & 0.883 [0.840, 0.919]     \\
\midrule
\multicolumn{5}{l}{\textit{Instruct regime (Aya native-speaker)}} \\
\midrule
KGW     & Mistral & 0.989 [0.984, 0.994]     & 0.255 [0.202, 0.306]     & 0.892 [0.875, 0.909]     \\
        & Gemma   & 0.990 [0.982, 0.997]     & 0.397 [0.377, 0.420]     & 0.967 [0.957, 0.977]     \\
        & Qwen    & 0.987 [0.977, 0.995]     & 0.300 [0.269, 0.332]     & 0.885 [0.846, 0.918]     \\
\cmidrule(lr){1-5}
Unigram & Mistral & 0.981 [0.971, 0.990]     & 0.240 [0.188, 0.291]     & 0.897 [0.869, 0.927]     \\
        & Gemma   & 0.996 [0.995, 0.998]     & 0.400 [0.378, 0.423]     & 0.965 [0.956, 0.974]     \\
        & Qwen    & 0.980 [0.965, 0.992]     & 0.297 [0.263, 0.331]     & 0.885 [0.838, 0.924]     \\
\cmidrule(lr){1-5}
SynthID & Mistral & 0.721 [0.513, 0.911]     & 0.149 [0.065, 0.230]     & 0.595 [0.453, 0.723]     \\
        & Gemma   & 0.994 [0.990, 0.997]     & 0.365 [0.344, 0.388]     & 0.931 [0.915, 0.946]     \\
        & Qwen    & 0.833 [0.694, 0.940]     & 0.219 [0.168, 0.266]     & 0.645 [0.579, 0.710]     \\
\cmidrule(lr){1-5}
EXPEdit & Mistral & 0.480 [0.300, 0.659]     & 0.082 [0.004, 0.157]     & 0.549 [0.422, 0.675]     \\
        & Gemma   & 0.840 [0.769, 0.912]     & 0.319 [0.300, 0.339]     & 0.742 [0.722, 0.762]     \\
        & Qwen    & 0.699 [0.553, 0.831]     & 0.174 [0.128, 0.217]     & 0.651 [0.588, 0.713]     \\
\cmidrule(lr){1-5}
DIP     & Mistral & 0.992 [0.984, 0.997]     & 0.259 [0.206, 0.312]     & 0.937 [0.910, 0.961]     \\
        & Gemma   & 0.997 [0.995, 0.998]     & 0.403 [0.381, 0.425]     & 0.982 [0.973, 0.990]     \\
        & Qwen    & 0.994 [0.991, 0.997]     & 0.318 [0.287, 0.349]     & 0.925 [0.903, 0.944]     \\
\cmidrule(lr){1-5}
XSIR    & Mistral & 0.991 [0.987, 0.994]     & 0.258 [0.203, 0.314]     & 0.928 [0.911, 0.944]     \\
        & Gemma   & 0.995 [0.990, 0.998]     & 0.396 [0.374, 0.419]     & 0.979 [0.974, 0.984]     \\
        & Qwen    & 0.991 [0.987, 0.995]     & 0.311 [0.280, 0.342]     & 0.902 [0.880, 0.926]     \\
\bottomrule
\end{tabular}
\caption{95\% bootstrap intervals for the per-cell quality means of Table~\ref{tab:q_5_2_summary}, subset \texttt{all}. Each entry is the mean preservation score over the eleven languages with a percentile interval from $B = 1000$ resamples of the 11-language vector, as for the detection means of Table~\ref{tab:detection_summary}. The interval describes spread across languages, not sampling error within a language.}
\label{tab:q_5_2_ci}
\end{table}

Table~\ref{tab:q_5_2_summary} reports the per-(scheme, generator, regime) quality-fairness summary that the heatmaps in Figure~\ref{fig:quality} and the per-scheme decomposition in Figure~\ref{fig:ge2-decomposition} aggregate over. Three quantities are recoverable here that are not directly readable from those figures: the per-cell minimum-language identity (ISO 639-3 superscript on each mean--min pair), the four-fifths disparate-impact count DI$<$.8 against the per-cell maximum, and the GE$_2$ total with its between-family share at the (scheme, generator) granularity rather than the per-scheme generator-averaged values plotted in Figure~\ref{fig:ge2-decomposition}. Table~\ref{tab:q_5_2_ci} gives 95\% bootstrap intervals for each mean. The BERTScore DI$<$.8 and GE$_2$ columns are omitted: raw F1 has insufficient dynamic range to register either statistic, and the rescaled F1 used elsewhere is unbounded below and produces pathological ratios under both the four-fifths anchor and the squared-coefficient-of-variation normalization.

The cell-level numbers support three observations that the main-text aggregates state at the scheme level. First, the floor-language column shows that the worst-preserved language within a cell is not constant across (scheme, generator): Arabic, Hindi, Turkish, Vietnamese, and Japanese each appear as floor language in at least four cells across the grid, while English never does. Second, the DI$<$.8 column localizes the distortion-free collapse of \S\ref{sec:quality} to its cell-level extent: all six SynthID and EXPEdit base cells flag $\geq 6$ languages on MAUVE, while all six DIP base cells flag zero. Third, the per-cell GE$_2$ values expose within-scheme generator variation that the per-scheme bars in Figure~\ref{fig:ge2-decomposition} smooth over, the largest case being EXPEdit on MAUVE in the base regime, where Qwen (0.874) sits substantially above Mistral (0.536) and Gemma (0.574) despite all three being labeled ``EXPEdit base'' in the headline plot.

\subsection{PPL Preservation Under mGPT Cross-Reference}
\label{app:ppl-mgpt}

The cross-scheme ranking is invariant to reference choice. Per-scheme means across the 33 base cells: XSIR (0.93 mGPT, 0.91 XGLM), Unigram (0.92, 0.89), KGW (0.88, 0.87), DIP (0.81, 0.80), SynthID (0.33, 0.29), EXPEdit (0.22, 0.19). Per-cell agreement is close throughout the grid. The Hindi-on-PPL pattern flagged in \S\ref{sec:quality} reproduces under mGPT: KGW, Unigram, and XSIR all preserve Hindi at $\geq 0.65$ under both references despite weaker BERTScore and MAUVE preservation in the same cells. The largest single-cell shift between references is SynthID-Gemma on Hindi (0.27 XGLM $\to$ 0.51 mGPT); the second-largest is XSIR-Qwen on Hindi (0.75 $\to$ 0.98). Both shifts are toward higher preservation under mGPT, so the cross-reference comparison does not weaken the per-scheme conclusions of \S\ref{sec:quality}. Two-reference agreement supports the \S\ref{sec:framework} design choice of treating PPL as a single paradigm with one primary and one cross-reference reading.

\begin{figure}[htbp]
  \centering
  \begin{subfigure}[htbp]{0.48\linewidth}
    \centering
    \includegraphics[width=\linewidth]{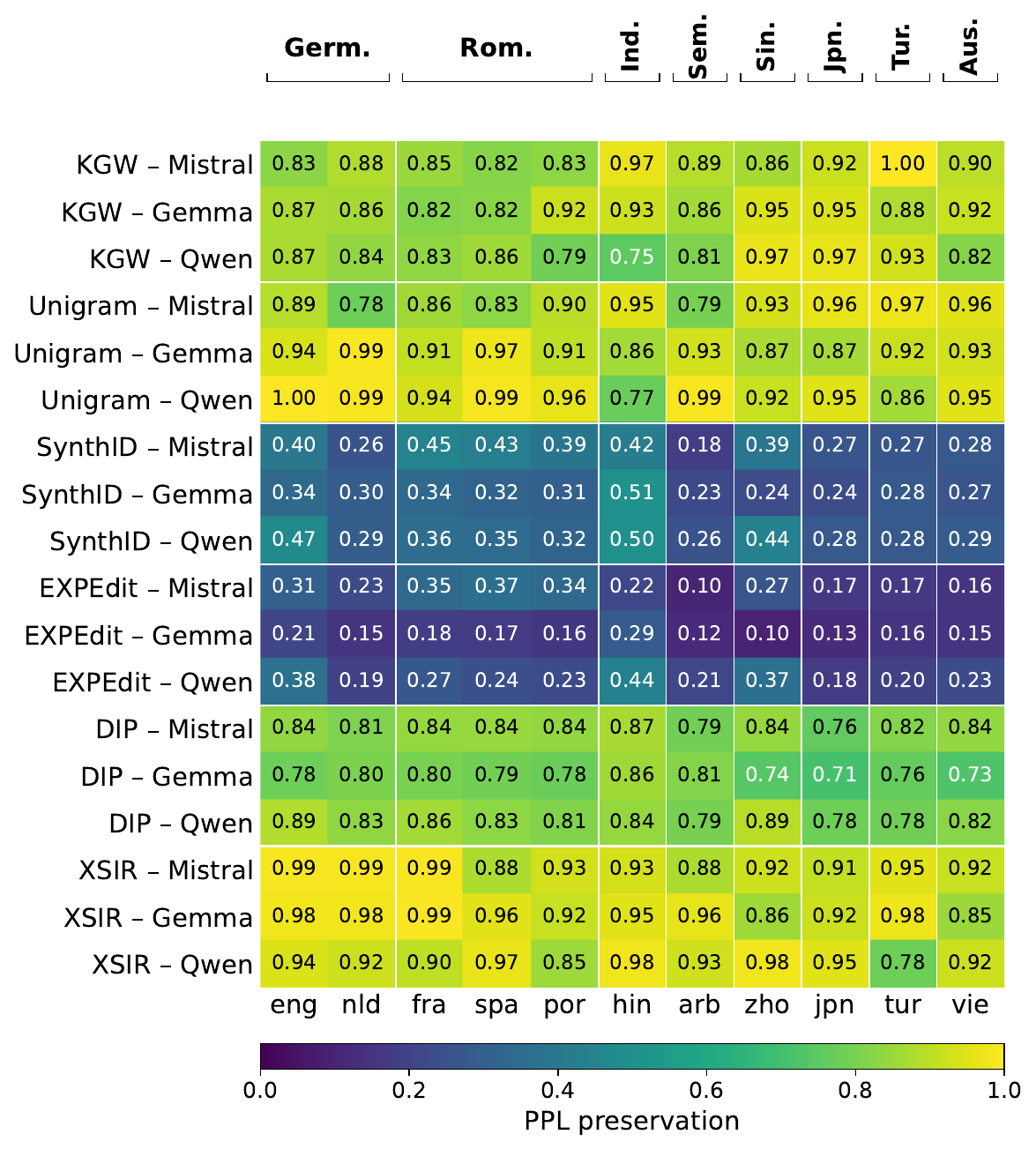}
    \caption{Base-FLORES regime.}
    \label{fig:ppl-mgpt-base}
  \end{subfigure}
  \hfill
  \begin{subfigure}[htbp]{0.48\linewidth}
    \centering
    \includegraphics[width=\linewidth]{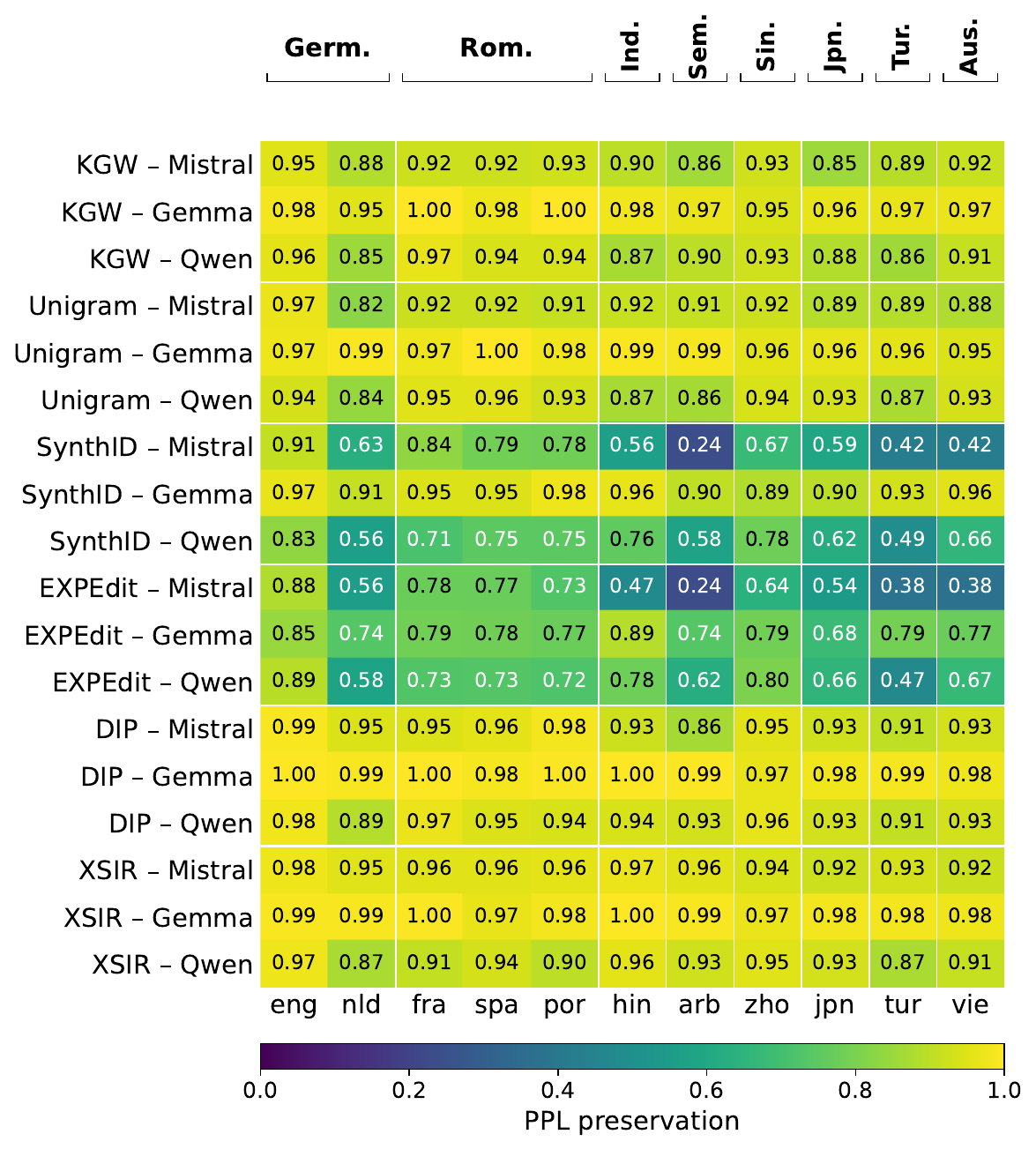}
    \caption{Instruct-Aya regime.}
    \label{fig:ppl-mgpt-instruct}
  \end{subfigure}
  \caption{PPL preservation under mGPT-1.3B cross-reference. Format and color scale match Figure~\ref{fig:quality}.}
  \label{fig:ppl-mgpt}
\end{figure}

\subsection{On-Target Vs.\ All-Subset Stratification}
\label{app:qstratify}
\begin{figure}[htbp]
  \centering
  \includegraphics[width=\linewidth]{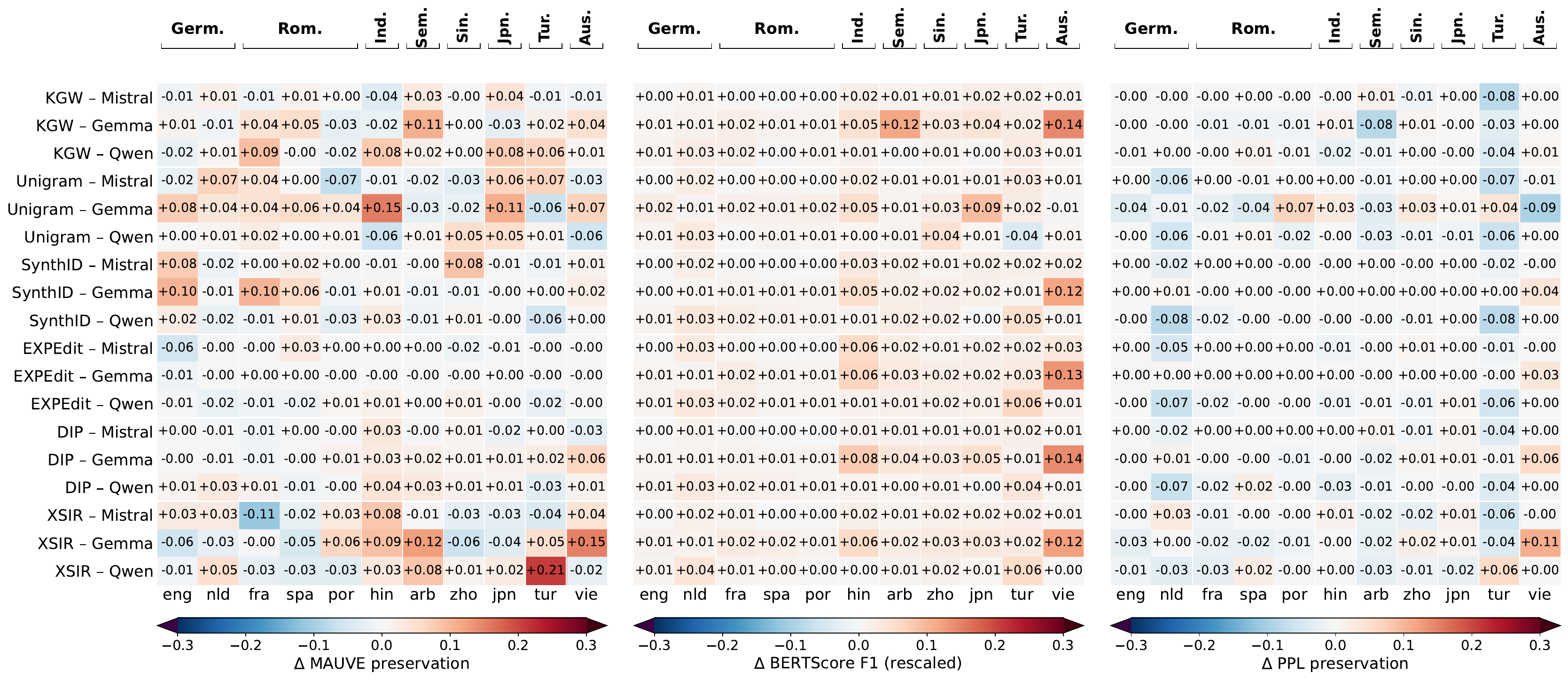}
  \vspace{0.5em}
  \includegraphics[width=\linewidth]{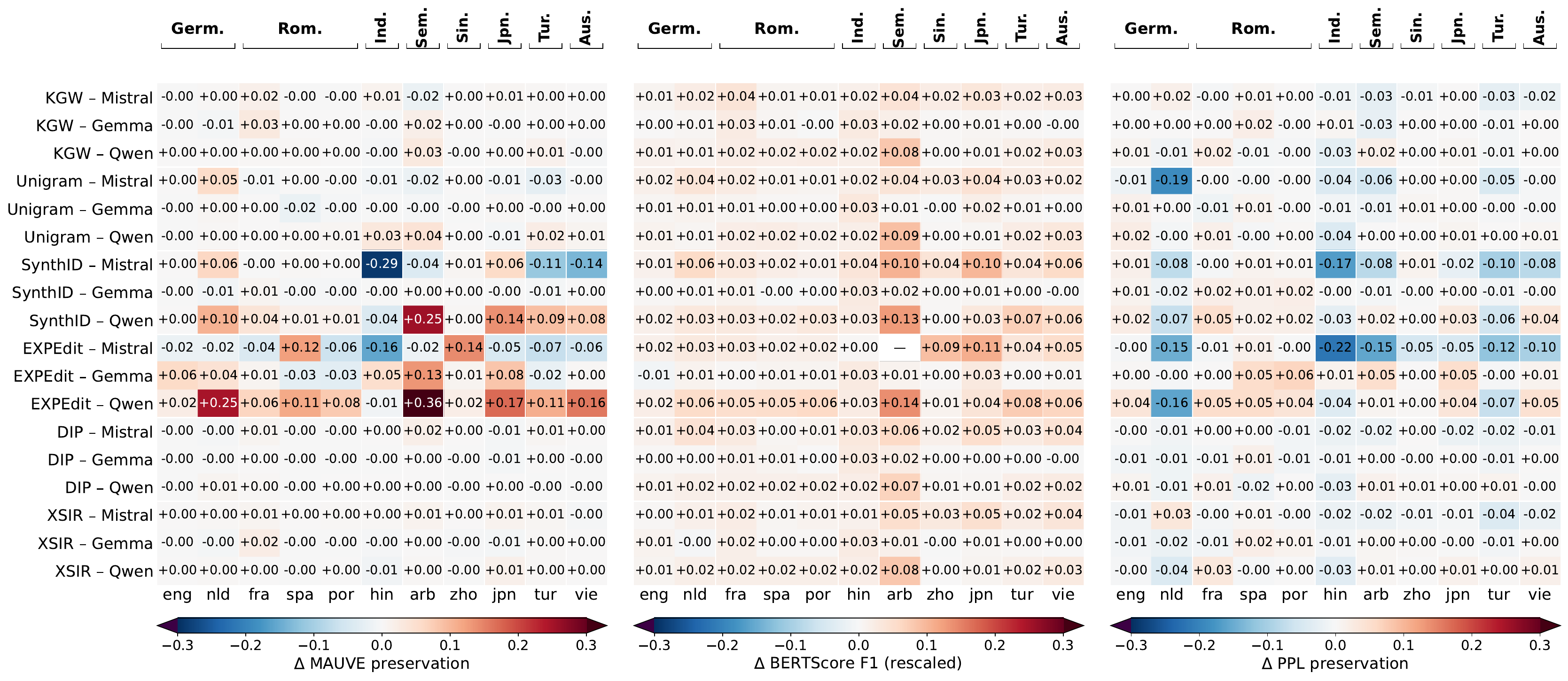}
  \caption{On-target minus all-subset preservation, base-FLORES regime (top) and instruct-Aya regime (bottom). Each cell is $\Delta = \text{preservation}_{\text{on-target}} - \text{preservation}_{\text{all}}$ for the indicated paradigm; rows are (scheme, generator), columns are languages. Positive (red): on-target restriction raises preservation; negative (blue): on-target restriction lowers it. Color scale fixed at $[-0.3, +0.3]$; values outside this range saturate. The EXPEdit-Mistral, Arabic, BERTScore F1 entry in the instruct panel is omitted (---) because the on-target sample size falls below the $n \geq 100$ paired-set floor.}
  \label{fig:qstrat}
\end{figure}

The headline quality reporting in \S\ref{sec:quality} is computed on subset \texttt{=all} per the pre-registration. Figure~\ref{fig:qstrat} reports the per-cell shift $\Delta = \text{preservation}_{\text{on-target}} - \text{preservation}_{\text{all}}$ across the three paradigms in both regimes.

In base, shifts fall within $\pm 0.05$ for the vast majority of cells, with isolated MAUVE exceptions on Unigram-Gemma, XSIR-Gemma, and XSIR-Qwen that do not displace the cross-scheme rankings of \S\ref{sec:quality}. In instruct, shifts concentrate on the most-distorting schemes (SynthID, EXPEdit) and the lowest-adherence cells flagged in Table~\ref{tab:p1-adherence} (Mistral and Qwen on Arabic, Hindi, Turkish, Vietnamese). Two opposing patterns recur on these cells: MAUVE shifts are large and positive (on-target restriction raises apparent preservation, indicating that off-target generations are distributionally far from baseline), while PPL preservation shifts are predominantly negative (on-target generations are distributionally closer to baseline but exhibit lower fluency under the held-out reference). The bidirectionality reflects that off-target removal trades distributional similarity against reference-perplexity fluency in opposite directions on these cells, consistent with the green-list/sampling-time mechanism split. The four logit-bias schemes show shifts within $\pm 0.05$ on the great majority of entries in both regimes.

With the single exception of the EXPEdit-Mistral, Arabic, BERTScore entry (omitted for falling below the $n \geq 100$ paired-set floor), the conclusions of \S\ref{sec:quality} survive on-target restriction: SynthID and EXPEdit remain the most distorting, the cross-paradigm disagreement pattern persists, and the MAUVE between-family shares barely move (medians 63.4\% $\to$ 63.8\% base, 90.0\% $\to$ 90.4\% instruct). Table~\ref{tab:btw_all_vs_ontarget} reports the per-cell shares behind these medians and behind the detection medians of \S\ref{sec:findings_detection}.

\begin{table}[htbp]
\centering
\scriptsize
\setlength{\tabcolsep}{3.5pt}
\begin{tabular}{l l cc cc cc cc}
\toprule
& & \multicolumn{4}{c}{Base regime (FLORES+)} & \multicolumn{4}{c}{Instruct regime (Aya)} \\
\cmidrule(lr){3-6} \cmidrule(lr){7-10}
& & \multicolumn{2}{c}{Detection} & \multicolumn{2}{c}{MAUVE} & \multicolumn{2}{c}{Detection} & \multicolumn{2}{c}{MAUVE} \\
\cmidrule(lr){3-4} \cmidrule(lr){5-6} \cmidrule(lr){7-8} \cmidrule(lr){9-10}
Scheme & Gen. & all & on-target & all & on-target & all & on-target & all & on-target \\
\midrule
KGW     & Mistral & (83.6)  & (87.5)  & (86.0)  & (87.6)  & 95.3    & 97.4    & (58.9)  & (98.3)  \\
        & Gemma   & (72.4)  & (74.2)  & 98.9    & 98.7    & 82.4    & 79.8    & (65.3)  & (88.3)  \\
        & Qwen    & (85.1)  & (96.1)  & 95.6    & 99.9    & 78.6    & 79.3    & (90.8)  & (92.2)  \\
\cmidrule(lr){1-10}
Unigram & Mistral & 61.3    & 59.8    & 49.5    & 64.7    & 99.8    & 99.9    & (40.3)  & (90.9)  \\
        & Gemma   & ---     & (86.9)  & 67.8    & 65.1    & 81.2    & 93.0    & (54.1)  & (39.3)  \\
        & Qwen    & 93.8    & 99.4    & 97.2    & 97.1    & 91.7    & 93.8    & (97.7)  & (95.3)  \\
\cmidrule(lr){1-10}
SynthID & Mistral & (100.0) & (100.0) & 59.0    & 55.6    & 77.7    & 82.6    & 99.6    & 99.9    \\
        & Gemma   & (100.0) & (100.0) & 43.0    & 40.5    & 79.7    & 77.7    & (50.0)  & (62.9)  \\
        & Qwen    & (97.5)  & (57.1)  & 53.0    & 52.3    & 91.9    & 95.0    & 97.1    & 99.5    \\
\cmidrule(lr){1-10}
EXPEdit & Mistral & (88.0)  & ---     & 57.1    & 64.2    & 80.0    & 81.5    & 86.6    & 87.2    \\
        & Gemma   & (100.0) & (100.0) & 43.4    & 42.9    & 84.2    & 81.4    & 90.0    & 75.6    \\
        & Qwen    & (99.2)  & (99.3)  & 47.9    & 45.7    & 84.4    & 82.1    & 80.6    & 90.4    \\
\cmidrule(lr){1-10}
DIP     & Mistral & (92.6)  & (96.7)  & (95.6)  & (77.4)  & 99.5    & 99.7    & (96.6)  & (98.6)  \\
        & Gemma   & (99.5)  & (96.2)  & (94.7)  & (95.2)  & 91.6    & 89.6    & (52.2)  & (78.8)  \\
        & Qwen    & 94.7    & 99.4    & (81.6)  & (92.8)  & 99.6    & 99.7    & (28.0)  & (38.1)  \\
\cmidrule(lr){1-10}
XSIR    & Mistral & 91.5    & 93.1    & 86.6    & 54.6    & 91.8    & 90.4    & (91.9)  & (97.1)  \\
        & Gemma   & 96.0    & 94.6    & 80.9    & 63.5    & 76.5    & 83.7    & (31.1)  & (80.8)  \\
        & Qwen    & 98.9    & 98.9    & 89.9    & 92.7    & 99.9    & 100.0   & (90.6)  & (96.5)  \\
\midrule
\multicolumn{2}{l}{Median (cells above floor)} & 94.2 & 96.7 & 63.4 & 63.8 & 88.0 & 90.0 & 90.0 & 90.4 \\
\bottomrule
\end{tabular}
\caption{Per-cell between-family share (\%) of GE$_2$ under the typological partition, subset \texttt{all} against the on-target subset, for detection TPR at $\tau_g$ ($\alpha = 0.01$; on-target thresholds recalibrated on each cell's on-target pool) and for MAUVE preservation. Every cell retains all eleven languages in both subsets. Shares of cells with GE$_2$ below $10^{-3}$ are given in parentheses and are excluded from the medians, which are taken over the cells above the floor within each subset; --- marks cells whose share is undefined (a per-language mean of zero, or GE$_2$ exactly zero). The \texttt{all} columns repeat the Btw.\,\% columns of Tables~\ref{tab:detection_summary} and \ref{tab:q_5_2_summary}.}
\label{tab:btw_all_vs_ontarget}
\end{table}

\section{Joint Detection-Quality: Full Paradigm Grid}\label{app:joint}
\begin{figure}[htbp]
\centering
\includegraphics[width=\textwidth]{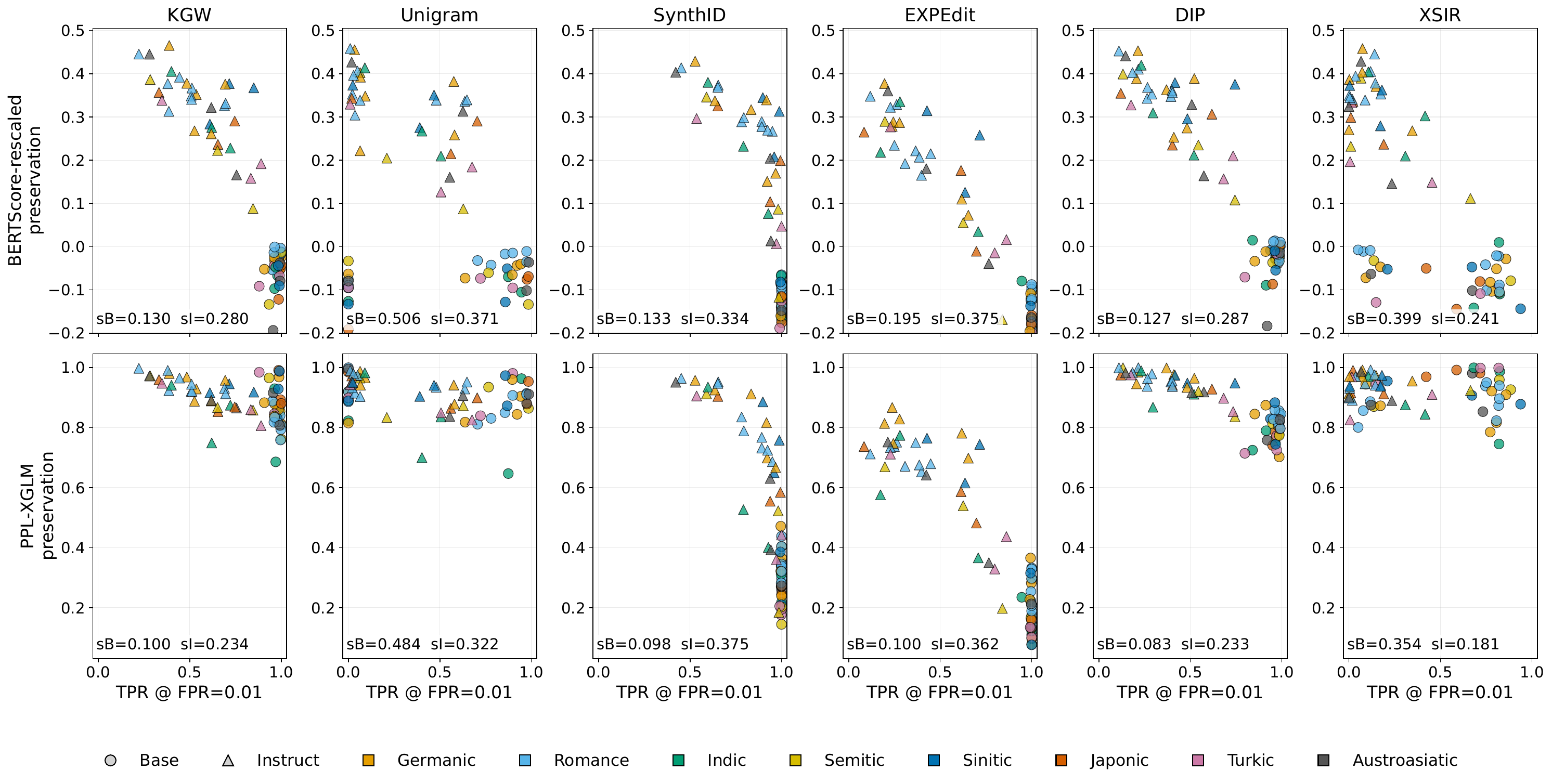}
\caption{Joint per-language detection--quality landscape, BERTScore-rescaled (top) and PPL-XGLM (bottom) paradigms. Format matches Figure~\ref{fig:pareto-headline}: markers are per-language $(\mathrm{TPR}\,@\,\tau_g, \mathrm{quality})$ pairs at $\alpha{=}0.01$, $n{=}33$ per regime (11 languages $\times$ 3 generators); circles are base-FLORES, triangles are instruct-Aya; color encodes typological family; $s_B$, $s_I$ are mean pairwise Euclidean distance within each regime.}
\label{fig:pareto-appendix}
\end{figure}

Figure~\ref{fig:pareto-appendix} reports the BERTScore-rescaled and PPL-XGLM companion panels to Figure~\ref{fig:pareto-headline}. The topology of \S\ref{sec:joint} carries through on both paradigms. Unigram and XSIR retain the highest base-regime spread on every paradigm (Unigram $s_B \in [0.484, 0.551]$; XSIR $s_B \in [0.354, 0.399]$), with the four remaining schemes at $s_B \le 0.195$. EXPEdit, SynthID, and Unigram lead the instruct-regime spread on every paradigm, and XSIR carries the lowest instruct spread on every paradigm ($s_I \in [0.159, 0.241]$). The opposite-corner low-spread topologies of \S\ref{sec:joint} (DIP top-right, EXPEdit bottom-right) reproduce on the BERTScore and PPL panels: DIP base $s_B \le 0.127$ across paradigms, EXPEdit base $s_B \le 0.195$. The headline finding, that joint disparity is regime-shaped rather than scheme-uniform, is therefore not specific to MAUVE.

\section{Mechanism Regression}
\label{app:mechanism}

The decompositions of \S\ref{sec:findings_detection}-\S\ref{sec:joint} establish that detection disparity is typology-structured but do not identify which structural variable does the work. This appendix reports per-language OLS fits of detection TPR and quality preservation on four candidate covariates: tokenizer fertility (tokens per FLORES-devtest sentence), unwatermarked-generation surprisal under XGLM-7.5B, a Latin-script indicator, and language-adherence rate (NWM\% from Table~\ref{tab:p1-adherence}). Tables~\ref{tab:m_5_4_r2_tpr} and \ref{tab:m_5_4_r2_mauve} report per-regime single-covariate $R^2$ over 33 (language, generator) points; \S\ref{app:mechanism-paradigm} repeats the quality-side analysis on BERTScore-rescaled and PPL-XGLM as paradigm-stability checks on the MAUVE headline.

\textit{Detection and quality have different mechanisms.} In base, adherence is the dominant detection predictor for KGW, SynthID, and DIP ($R^2$ between $0.41^{**}$ and $0.60^{**}$); fertility is dominant only for EXPEdit ($0.83^{**}$), though it is also significant for DIP ($0.24^{**}$). On quality, fertility predicts the green-list schemes (KGW $0.39^{**}$, Unigram $0.25^{**}$, DIP $0.23^{**}$), and adherence registers only for the distortion-free schemes under instruct. The covariate that drives detectability is not the one that drives preservation.

\textit{Instruction tuning collapses detection-side structure.} Detection $R^2$ falls below significance for five of six schemes under instruct (XSIR adherence $0.29^{**}$ excepted), consistent with the calibration-driven instruct collapse of \S\ref{sec:findings_detection}. Quality-side fertility persists for the green-list family. Latin script and NWM surprisal clear $p<0.05$ in 6 of 48 cells combined; both correlate with the typological partition but neither carries enough non-collinear signal at $n{=}33$ to compete. Pooling across regimes (results omitted) dilutes the regime-specific detection mechanisms while preserving the regime-stable quality-side fertility effect, corroborating the asymmetry.

\begin{table}[htbp]
\centering
\scriptsize
\begin{tabular}{lcccccc}
\toprule
Covariate & KGW & Unigram & SynthID & EXPEdit & DIP & XSIR \\
\midrule
\multicolumn{7}{l}{\textit{Base regime (FLORES+ continuation, $n=33$)}} \\
\midrule
Fertility (tokens/sent.) & 0.016 & 0.052 & 0.008 & 0.834$^{**}$ & 0.238$^{**}$ & 0.001 \\
NWM surprisal (XGLM)     & 0.066 & 0.074 & 0.058 & 0.261$^{**}$ & 0.156$^{*}$  & 0.003 \\
Latin script             & 0.029 & 0.001 & 0.005 & 0.051        & 0.009        & 0.019 \\
NWM adherence            & 0.600$^{**}$ & 0.006 & 0.415$^{**}$ & 0.001 & 0.461$^{**}$ & 0.091 \\
\midrule
\multicolumn{7}{l}{\textit{Instruct regime (Aya native-speaker, $n=33$)}} \\
\midrule
Fertility (tokens/sent.) & 0.017 & 0.046 & 0.009 & 0.001 & 0.000 & 0.107 \\
NWM surprisal (XGLM)     & 0.024 & 0.003 & 0.009 & 0.040 & 0.017 & 0.011 \\
Latin script             & 0.046 & 0.014 & 0.035 & 0.049 & 0.040 & 0.058 \\
NWM adherence            & 0.000 & 0.000 & 0.003 & 0.047 & 0.000 & 0.291$^{**}$ \\
\bottomrule
\end{tabular}
\caption{Single-covariate $R^2$ for per-language detection TPR at $\tau_g$,
$\alpha{=}0.01$, subset all. Each fit regresses the per-language outcome
vector on a single covariate over 33 (language, generator) points.
Stars from the slope $p$-value: $^{*}p<0.05$, $^{**}p<0.01$.}
\label{tab:m_5_4_r2_tpr}
\end{table}

\begin{table}[htbp]
\centering
\scriptsize
\begin{tabular}{lcccccc}
\toprule
Covariate & KGW & Unigram & SynthID & EXPEdit & DIP & XSIR \\
\midrule
\multicolumn{7}{l}{\textit{Base regime (FLORES+ continuation, $n=33$)}} \\
\midrule
Fertility (tokens/sent.) & 0.392$^{**}$ & 0.249$^{**}$ & 0.027 & 0.005 & 0.234$^{**}$ & 0.093 \\
NWM surprisal (XGLM)     & 0.026        & 0.023        & 0.018 & 0.001 & 0.007        & 0.009 \\
Latin script             & 0.136$^{*}$  & 0.036        & 0.062 & 0.052 & 0.083        & 0.000 \\
NWM adherence            & 0.031        & 0.007        & 0.101 & 0.074 & 0.219$^{**}$ & 0.072 \\
\midrule
\multicolumn{7}{l}{\textit{Instruct regime (Aya native-speaker, $n=33$)}} \\
\midrule
Fertility (tokens/sent.) & 0.346$^{**}$ & 0.464$^{**}$ & 0.027        & 0.016        & 0.019 & 0.168$^{*}$ \\
NWM surprisal (XGLM)     & 0.008        & 0.152$^{*}$  & 0.022        & 0.052        & 0.005 & 0.002 \\
Latin script             & 0.122$^{*}$  & 0.051        & 0.018        & 0.041        & 0.052 & 0.082 \\
NWM adherence            & 0.076        & 0.099        & 0.153$^{*}$  & 0.230$^{**}$ & 0.053 & 0.047 \\
\bottomrule
\end{tabular}
\caption{Single-covariate $R^2$ for per-language MAUVE preservation. Each
fit regresses the per-language outcome vector on a single covariate over
33 (language, generator) points. Stars from the slope $p$-value:
$^{*}p<0.05$, $^{**}p<0.01$. BERTScore-rescaled and PPL-XGLM
paradigm-stability tables in \S\ref{app:mechanism-paradigm}.}
\label{tab:m_5_4_r2_mauve}
\end{table}

\subsection{Paradigm-Stability Check}
\label{app:mechanism-paradigm}

Tables~\ref{tab:m_5_4_r2_bertscore_rescaled} and \ref{tab:m_5_4_r2_ppl_xglm} repeat the single-covariate quality-side regression with BERTScore-rescaled and PPL-XGLM as the outcome. PPL-XGLM corroborates and strengthens the MAUVE story: fertility is the dominant predictor for the green-list family on both regimes, with the largest $R^2$ values in Table~\ref{tab:m_5_4_r2_ppl_xglm} appearing under instruct (KGW $0.446^{**}$, Unigram $0.559^{**}$, XSIR $0.340^{**}$). BERTScore-rescaled diverges: adherence, not fertility, is the dominant predictor in base for five of six schemes ($R^2$ between $0.188^{*}$ and $0.378^{**}$). The divergence is consistent with the narrow-dynamic-range observation of Appendix\ref{app:quality-summary}: BERTScore-rescaled F1 is sensitive to the per-language LID-baseline that goes into rescaling, so its single-covariate $R^2$ captures adherence rather than the fertility-driven distributional shift that MAUVE and PPL-XGLM register. The fertility-on-green-list mechanism is therefore paradigm-stable across the two paradigms that measure distributional or reference-likelihood divergence directly.

\begin{table}[htbp]
\centering
\scriptsize
\begin{tabular}{lcccccc}
\toprule
Covariate & KGW & Unigram & SynthID & EXPEdit & DIP & XSIR \\
\midrule
\multicolumn{7}{l}{\textit{Base regime ($n=33$)}} \\
\midrule
Fertility (tokens/sent.) & 0.000        & 0.013        & 0.024        & 0.054        & 0.039        & 0.095 \\
NWM surprisal (XGLM)     & 0.036        & 0.016        & 0.000        & 0.032        & 0.039        & 0.094 \\
Latin script             & 0.068        & 0.207$^{**}$ & 0.004        & 0.022        & 0.003        & 0.008 \\
NWM adherence            & 0.239$^{**}$ & 0.004        & 0.378$^{**}$ & 0.259$^{**}$ & 0.298$^{**}$ & 0.188$^{*}$ \\
\midrule
\multicolumn{7}{l}{\textit{Instruct regime ($n=33$)}} \\
\midrule
Fertility (tokens/sent.) & 0.072 & 0.069 & 0.043        & 0.026        & 0.047        & 0.051 \\
NWM surprisal (XGLM)     & 0.034 & 0.034 & 0.030        & 0.021        & 0.037        & 0.026 \\
Latin script             & 0.046 & 0.045 & 0.042        & 0.037        & 0.044        & 0.056 \\
NWM adherence            & 0.096 & 0.095 & 0.119$^{*}$  & 0.125$^{*}$  & 0.123$^{*}$  & 0.110 \\
\bottomrule
\end{tabular}
\caption{Single-covariate $R^2$ for BERTScore-rescaled preservation. Stars from the slope $p$-value: $^{*}p<0.05$, $^{**}p<0.01$.}
\label{tab:m_5_4_r2_bertscore_rescaled}
\end{table}

\begin{table}[htbp]
\centering
\scriptsize
\begin{tabular}{lcccccc}
\toprule
Covariate & KGW & Unigram & SynthID & EXPEdit & DIP & XSIR \\
\midrule
\multicolumn{7}{l}{\textit{Base regime ($n=33$)}} \\
\midrule
Fertility (tokens/sent.) & 0.191$^{*}$  & 0.327$^{**}$ & 0.013        & 0.000        & 0.121$^{*}$  & 0.148$^{*}$ \\
NWM surprisal (XGLM)     & 0.003        & 0.179$^{*}$  & 0.017        & 0.018        & 0.094        & 0.183$^{*}$ \\
Latin script             & 0.022        & 0.001        & 0.126$^{*}$  & 0.105        & 0.052        & 0.056 \\
NWM adherence            & 0.186$^{*}$  & 0.014        & 0.164$^{*}$  & 0.123$^{*}$  & 0.109        & 0.015 \\
\midrule
\multicolumn{7}{l}{\textit{Instruct regime ($n=33$)}} \\
\midrule
Fertility (tokens/sent.) & 0.446$^{**}$ & 0.559$^{**}$ & 0.119$^{*}$ & 0.070        & 0.225$^{**}$ & 0.340$^{**}$ \\
NWM surprisal (XGLM)     & 0.044        & 0.029        & 0.010       & 0.000        & 0.047        & 0.016 \\
Latin script             & 0.086        & 0.077        & 0.070       & 0.084        & 0.046        & 0.012 \\
NWM adherence            & 0.020        & 0.024        & 0.096       & 0.217$^{**}$ & 0.023        & 0.006 \\
\bottomrule
\end{tabular}
\caption{Single-covariate $R^2$ for PPL-XGLM preservation. Stars from the slope $p$-value: $^{*}p<0.05$, $^{**}p<0.01$.}
\label{tab:m_5_4_r2_ppl_xglm}

\end{table}

\subsection{Design-Axis Ranking and Mitigation Levers}\label{app:levers}

Four design axes could in principle account for the gaps documented in \S\ref{sec:findings}: (a) tokenizer vocabulary coverage, (b) the interaction of green-list partitioning with script, (c) the language distribution of pre-training, and (d) morphological richness, which affects how independent successive token choices are. The regressions above observe these axes through proxies, with fertility standing in for (a) and (d), the Latin-script indicator for (b), and adherence for (c). For detection the order is (c) $>$ (a) $>$ (b) $>$ (d). Adherence is the dominant predictor for three of six schemes in the base regime ($R^2$ between 0.41 and 0.60, Table~\ref{tab:m_5_4_r2_tpr}) and fertility is dominant only for EXPEdit (0.83). A model that is weak in a language drifts off-language and generates with less usable choice which the watermark needs to embed its signal. For quality the order is (a) $>$ (d) $>$ (c) $>$ (b). Fertility is the dominant predictor for the green-list schemes in both regimes (Tables~\ref{tab:m_5_4_r2_mauve} and \ref{tab:m_5_4_r2_ppl_xglm}), because a language cut into more pieces per sentence receives more interventions per sentence. Adherence registers mainly for the distortion-free schemes, and the script indicator is significant in only two MAUVE fits, both for KGW. Axes (a) and (d) cannot be separated at $n{=}33$, since Turkish is the only agglutinative language in the grid, but both surface through fertility, so the mitigation is the same.

The family partition captures more between-group variance than the script partition in every cell with non-trivial disparity (Table~\ref{tab:ge_partition_robustness}). Against its own exact null, a Latin-versus-non-Latin partition clears $p<.05$ in 1 of 18 instruct detection cells, against 8 of 18 for an Indo-European-versus-rest partition (Appendix~\ref{app:sensitivity}). The script share approaches the family share only in XSIR-Qwen, whose green list is built from cross-lingual semantic clusters, with the cell being the Hindi outlier of \S\ref{sec:findings_detection}, in which the two partitions coincide. We therefore read it as pointing to partition construction, a parameter the scheme designer controls, rather than to the writing system itself.

The axes map onto three different actors. Deployers can calibrate thresholds empirically, per language and per deployment pool, which closes the calibration-driven part of the detection gaps (Tables~\ref{tab:detection_calibration_gap} and \ref{tab:detection_on_target}) and leaves quality untouched. Scheme designers control green-list and partition construction and could make both fertility-aware, so that a language split into more tokens per sentence does not receive proportionally more interventions. Model developers control vocabulary allocation and pre-training coverage, where the uneven treatment of languages is already documented~\cite{rust-etal-2021-good, petrov2023language, ahia-etal-2023-languages}.

\subsection{Scale Pilot: Qwen2.5-72B}\label{app:scale}
To test the two scale predictions of \S\ref{discussion}, we ran Qwen2.5-72B~\cite{yang2024qwen2} on KGW, SynthID, and XSIR in the base regime, over all eleven languages at $n = 500$ matched pairs per cell, with the same prompts, seeds, decoding settings, and numerical precision as the Qwen2.5-7B cells. The two models share a tokenizer, so fertility is held fixed by construction and scale is the only intended difference. The instruct-regime cells were not run. Table~\ref{tab:scale_pilot} reports the 7B and 72B values side by side.

\begin{table}[htbp]
\centering
\small
\begin{tabular}{lccc}
\toprule
Scheme (base) & Mean TPR & Min TPR (language) & MAUVE mean \\
\midrule
KGW     & $0.965 \to 0.936\,[0.895,\,0.971]$ & $0.880 \to 0.794$ (tur $\to$ nld) & $0.889 \to 0.922$ \\
SynthID & $0.997 \to 0.997\,[0.993,\,0.999]$ & $0.990 \to 0.982$ (tur $\to$ tur) & $0.183 \to 0.386$ \\
XSIR    & $0.215 \to 0.175\,[0.074,\,0.329]$ & $0.050 \to 0.040$ (por $\to$ por) & $0.850 \to 0.902$ \\
\bottomrule
\end{tabular}
\caption{Scale pilot, Qwen2.5-7B $\to$ Qwen2.5-72B, base regime, subset \texttt{all}, $\alpha = 0.01$, global threshold $\tau_g$, $n = 500$ matched pairs per cell. Brackets are 95\% bootstrap CIs at 72B ($B = 1000$ resamples of the 11-language vector, as in Table~\ref{tab:detection_summary}). Min TPR names the floor language at each size.}
\label{tab:scale_pilot}
\end{table}

The max-min TPR spread narrows in one of three cells, the worst-language TPR moves by a median of $-1.0$~pp, and XSIR remains the scheme with the lowest mean TPR. Quality levels rose although the tokenizer is fixed (median MAUVE $+0.052$, with SynthID moving from 0.183 to 0.386). These are changes in level, so the pilot does not directly test whether cross-language quality disparity persists, but the size of the SynthID shift is a caution against the strong form of the fertility prediction, under which quality tracks fertility alone and should not move when the tokenizer is fixed. The between-family shares are 86.7\% for detection and 89.8\% for MAUVE against shape-matched null medians of 71.9\% and 73.9\%, with Holm-adjusted $p = .171$ and $.145$ over the two strata and at most three cells per stratum. We therefore treat scale behavior as a hypothesis instead of a finding.


\end{document}